%% file: main.tex
\documentclass[letterpaper,twocolumn,10pt]{article}
\PassOptionsToPackage{hyphens}{url}
\usepackage{usenix2019_v3}

\usepackage{graphicx}
\usepackage{amsmath,amssymb}
\usepackage{booktabs}
\usepackage{caption}
\makeatletter
\def\section{\@startsection {section}{1}{\z@}{-3.5ex plus-1ex minus
    -.2ex}{2.3ex plus.2ex}{\reset@font\large\bf\raggedright}}
\renewcommand\subsection{\@startsection{subsection}{2}{\z@}{-3.25ex\@plus -1ex \@minus -.2ex}{1.5ex \@plus .2ex}{\normalfont\large\bfseries\raggedright}}
\makeatother

\usepackage{amsfonts}
\usepackage{xspace}
\usepackage{makecell}
\usepackage{pifont}
\usepackage{algorithm}
\usepackage[noend]{algpseudocode}

\newcommand{\fig}[1]{Figure~\ref{#1}}
\newcommand{\sect}[1]{Section~\ref{#1}}
\newcommand{\tab}[1]{Table~\ref{#1}}
\newcommand{\algo}[1]{Algorithm~\ref{#1}}

\newcommand{\circled}[1]{\ding{\the\numexpr 181+#1\relax}}

\newcommand{\proposed}[0]{A2\-Op\-ti\-mizer\xspace}
\newcommand{\simulator}[0]{A2Sim\xspace}
\newcommand{\exploresim}{\textit{ex\-plore}\textsubscript{\textit{sim}}\xspace}
\newcommand{\benchmarksim}{\textit{bench\-mark}\textsubscript{\textit{sim}}\xspace}
\newcommand{\diagnosesim}{\textit{di\-ag\-nose}\textsubscript{\textit{sim}}\xspace}
\newcommand{\benchmarkgpu}{\textit{bench\-mark}\textsubscript{GPU}\xspace}
\newcommand{\codertestgpu}{\textit{coder\-test}\textsubscript{GPU}\xspace}
\newcommand{\diagnosegpu}{\textit{di\-ag\-nose}\textsubscript{GPU}\xspace}
\algrenewcommand\algorithmicrequire{\textbf{Input:}}
\algrenewcommand\algorithmicensure{\textbf{Output:}}
\algrenewcommand\algorithmiccomment[1]{\hfill\textcolor{gray}{$\triangleright$ #1}}

\newcommand{\logofile}{a2sys_logo}
\newcommand{\logoheight}{20pt}
\newcommand{\logodrop}{30pt}
\newcommand{\logodate}{\the\year-\the\month-\the\day}
\AddToHookNext{shipout/background}{%
  \put(\dimexpr 1in+\hoffset+\oddsidemargin\relax,
       -\dimexpr 1in+\voffset+\topmargin+\headheight+\headsep+\logodrop\relax){%
    \parbox[b]{\textwidth}{%
      \includegraphics[height=\logoheight]{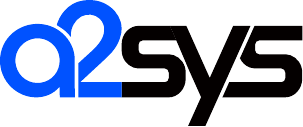}\hfill
      {\small\rmfamily\logodate}\par\vspace{3pt}%
      {\color{black!45}\hrule height 0.8pt}}}}

\begin{document}

\date{}
\title{\Large \bf \proposed{}: Accelerating Time-to-Efficiency of LLM Serving with Agents}
\author{%
{\rm Taekwon Lee$^{*\dagger}$, Kyoungho Jeun$^{*}$, Hyeseong Kim$^{*\dagger}$, Jiin Kim$^{*\dagger}$, Gunho Park$^{*}$,}\\
{\rm Byeongjun Shin$^{*\dagger}$, Seungjun Lee$^{*\dagger}$, Wontak Ryu$^{*}$, Seunghyun Lee$^{*}$, Deukwoong Woo$^{*}$,}\\
{\rm Woong Park$^{*}$, Minyoung Kim$^{*}$, Baeseong Park$^{*}$, Dongsoo Lee$^{*}$, Minsoo Rhu$^{*\dagger}$}\\[3pt]
{\rm $^{*}$a2sys, Inc.\qquad $^{\dagger}$KAIST}}

\maketitle

\input{tex/abstract}

\input{tex/intro}
\input{tex/background}
\input{tex/motivation}
\input{tex/proposed}
\input{tex/methodology}
\input{tex/evaluation}
\input{tex/related_work}
\input{tex/conclusion}

\bibliographystyle{plain}
\bibliography{refs}

\end{document}

%% file: tex/abstract.tex
\begin{abstract}
\label{sect:abstract}

Serving large language models (LLMs) efficiently is a recurring, time-critical optimization problem: each new model, hardware platform, and workload reopens the search for the serving configuration and code that maximize tokens per GPU under a latency objective. We call the time this search takes to reach a competitive operating point the \emph{time-to-efficiency} of a deployment. The search is increasingly delegated to AI coding agents, yet writing code is no longer what limits time-to-efficiency. Every optimization candidate must be benchmarked end to end on GPUs that could otherwise serve traffic, and new candidates for coding agents cannot start until these benchmarks finish. This paper presents \proposed{}, an Agentic AI based serving-stack Optimizer that shortens time-to-efficiency by removing most of this GPU benchmarking from the optimization loop. \proposed{} couples coding agents with \simulator{}, our performance simulator, which serves as a surrogate for the target LLM serving infrastructure. Most candidates are thus evaluated in simulation on CPUs, and GPUs are needed mainly for validation, which lets \proposed{} substantially reduce time-to-efficiency compared with a baseline agent loop for optimizing LLM serving.

\end{abstract}

%% file: tex/intro.tex
\section{Introduction}
\label{sect:introduction}

NeoCloud providers, i.e., cloud operators that specialize in GPU capacity~\cite{semianalysis_neocloud2024}, increasingly serve open-weight large language models (LLMs) on their own GPUs and sell the generated tokens through serverless endpoints, i.e., pay-per-token APIs that are offered directly~\cite{together_serverless,fireworks_serverless,baseten_modelapis} or through routing services~\cite{openrouter_quickstart,vercel_aigateway,hf_inference_providers}. When a competitive open-weight model is released, such a provider has two jobs, namely making the model available and making it efficient to serve. The first job is now routine, since major providers and widely used serving frameworks typically support new models on the day of release~\cite{fireworks_dayzero,vllm_day0,sglang_day0}, whereas the second can take months, because the serving stack must be tuned until it delivers a competitive number of tokens per GPU~\cite{vllm_example_1,vllm_example_2}. We call the optimization time needed to bring a new deployment to such a competitive, hardware-validated operating point its \emph{time-to-efficiency}. Until that point is reached, every token costs more than necessary, and because the open-weight LLM ecosystem continues to expand rapidly~\cite{open_llm_survey}, with competitive models now arriving just weeks apart, the next model often arrives before the current one has been fully optimized.

Whether a deployment reaches a competitive operating point depends on the serving software as much as on the hardware. On a fixed GPU platform, for example, software changes alone raised the per-GPU throughput of DeepSeek-R1 by 14$\times$~\cite{semianalysis_valuecapture2026}. Such gains take long to achieve, because they result from many interacting decisions about the \emph{configuration} of the serving stack, such as parallelism, batching, and speculative decoding, and about its \emph{code}, such as the engine's scheduler and its GPU kernels. Because the best combination of these decisions depends on the model, the hardware, the workload, and the service-level objective (SLO), a change in any of them reopens the search (\sect{subsect:serving_optimization}). This search is now increasingly being handed to AI coding agents, whose scope has widened from individual GPU kernels to the entire serving stack~\cite{openai_efficiency2026,meta_rea2026,liao2026kernelevolve,baseten_agentic_kernels2026,amd_hyperloom2026,cursor_multiagent2026,serverbench2026}.

In this paper, we begin by quantitatively characterizing, from the standpoint of a service provider that operates serverless LLM endpoints, what limits time-to-efficiency once AI agents themselves write the code that optimizes the serving stack. We find that time-to-efficiency is limited not by the agents' ability to write code but by the time spent on two key steps that run on real GPUs. The first is \emph{benchmarking}, which measures the end-to-end performance of an optimization candidate, i.e., a change that the agents have implemented, in order to decide whether to accept it. The second is \emph{diagnosis}, which analyzes where the serving system spends its time and how it uses its resources once a candidate has been accepted in order to identify the bottleneck that the next candidates should target. In a 24-hour agent-driven optimization run on an 8$\times$H100 server, benchmarking alone occupied the server for 18.8 hours, nearly 80\% of the run (\sect{subsect:challenge2}). Running more coding agents in parallel does not help, because the agents then produce candidates faster than the GPUs can benchmark them. Moreover, because every accepted change shifts the performance bottleneck, new candidates can be selected only after the updated serving system has been diagnosed again on GPUs, and each coding agent consequently remained idle for 89\% of its time on average (\sect{subsect:challenge_coder_orchestration}).

This paper presents \proposed{} (\textbf{A}gentic \textbf{A}I based serving-stack \textbf{Optimizer}), an agentic system that co-optimizes the configuration and code of an LLM serving stack end to end, so that a service provider can bring every newly released open-weight model to a competitive operating point within a fraction of today's time-to-efficiency. \proposed{} is designed to alleviate the two challenges identified above by co-designing the coding agents with an \emph{LLM serving simulator} in a closed loop, where each resolves the other's limitation. The agents can change the serving stack quickly but must wait for GPUs to benchmark and diagnose every change, whereas existing LLM serving simulators predict the performance that a serving system would achieve on its GPUs without actually running it~\cite{vidur,meta_inference_config2026,aiconfigurator_paper2026,dynosim2026} but assume a \emph{fixed} implementation. Our key insight is that a simulator can take the place of most GPU benchmarking and diagnosis, even while the agents keep rewriting the serving stack's code, if every code change is fed back into the simulator and its predictions drive both candidate acceptance and task assignment. To the best of our knowledge, this paper is the first to explore whether and how a simulator can take this role in an agent-driven serving-stack optimization loop, especially in the context of production serverless LLM endpoints.

\proposed{}, which we use in production to optimize our serverless LLM endpoints (\sect{sect:casestudy}), employs an \emph{orchestrator} agent and parallel \emph{coder} subagents, which work with our event-driven LLM serving simulator (henceforth \simulator{}) that acts as a digital twin of the serving systems behind our endpoints. The coders feed every code change back into \simulator{}, either as new kernel latency profiles or as a matching change to the code of the simulator, and \proposed{} uses the predictions of \simulator{} in three ways. First, before any code is written, a \emph{simulation-based serving configuration space exploration} selects a starting configuration. Second, \proposed{} \emph{benchmarks candidates in simulation} and benchmarks the accepted candidates on GPUs only in batches, using the measurements to recalibrate \simulator{}. Third, \proposed{} \emph{diagnoses the serving system in simulation} after every accepted candidate, which lets it \emph{orchestrate coders asynchronously}, so that a new candidate starts as soon as an earlier one receives its verdict. GPUs nevertheless remain the ground truth, since correctness is always verified on them and candidates outside \simulator's modeling scope are benchmarked on them directly.

We evaluate \proposed by optimizing vLLM-based serving of GLM-5.3-Flash,
DeepSeek-V4.1-Flash, and Qwen3.8-Flash-Next under AgentX, an open agentic-coding workload~\cite{agentx},
and GPT-OSS-120B under ShareGPT, a chat workload, on an 8$\times$H100 server. Compared with an agent loop that benchmarks every candidate on GPUs,
\proposed{} shortens time-to-efficiency by 1.1--15.1$\times$ (a geometric mean of 3.9$\times$), where we measure time-to-efficiency as the time needed to reach the highest per-GPU throughput that this loop attains within 24 hours. When both run for the full 24 hours,
\proposed{} also ends at a 1.06--1.11$\times$ higher per-GPU throughput and a 1.04--1.38$\times$ higher interactivity, i.e., the rate at which each user receives tokens.

%% file: tex/background.tex
\section{Background and Motivation}
\label{sect:background}\label{sect:motivation}

\subsection{LLM Serving Performance and Its Optimization Space}
\label{subsect:pareto_frontier}\label{subsect:serving_optimization}

{\bf Throughput--interactivity frontier.}
An LLM serves a request in two phases: \emph{prefill} processes the prompt and produces the first output token, and \emph{decode} generates the remaining tokens one at a time. Users experience the two phases as the \emph{time to first token} (TTFT) and the \emph{time per output token} (TPOT), and a service states its \emph{service-level objectives} (SLOs) as bounds on both. MLPerf Inference, for example, requires a 99th-percentile TTFT of 450\,ms and TPOT of 40\,ms (at least 25 tokens per second) in its interactive Llama~2 70B scenario~\cite{mlperf_llm_v5}. The provider's concern is \emph{throughput}, the tokens generated per second per GPU, which sets the cost per token (GPU cost per hour divided by tokens per hour).

\begin{figure}[t]
  \centering
  \includegraphics[width=0.95\columnwidth]{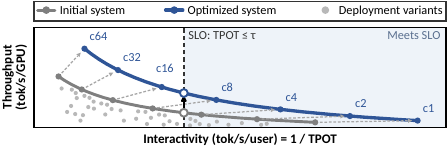}
    \vspace{0.3em}
  \captionsetup{font=small}
  \caption{The throughput--interactivity Pareto frontier (schematic). Each grey dot is one serving configuration measured at one concurrency level, and c$N$ marks a concurrency level of $N$. Optimization pushes the frontier toward the upper right (dashed arrows).}
  \label{fig:pareto_frontier}
  \vspace{1em}
\end{figure}

The latency that users want and the throughput that the provider wants cannot both be maximized, because batching more concurrent requests onto the same GPUs raises throughput but lengthens every decode iteration, so each user's stream slows down. \fig{fig:pareto_frontier} plots this trade-off in the form that NeoCloud providers, GPU vendors, open benchmark suites, and developers of serving engines now use~\cite{crusoe_inferencemax2025,nvidia_codesign2026,nvidia_dynamo_moe2025,inferencex,vllm_inferencemax2025}, namely throughput per GPU against \emph{interactivity}, the per-user generation rate (tokens/s/user), i.e., the reciprocal of TPOT. Each serving configuration is measured at several \emph{concurrency levels}, where a concurrency level is the number of requests that are kept in progress at the same time during a measurement, and each level yields one point on the plot, which we refer to as an \emph{operating point}. \fig{fig:pareto_frontier} labels the point that is measured at a concurrency level of $N$ as c$N$. Raising the concurrency level, e.g., from c1 to c64, moves a configuration's point toward the upper left, and the upper envelope of all points is the \emph{Pareto frontier}, i.e., the highest throughput attainable at each level of interactivity. On this plot, an SLO on TPOT appears as a vertical line, and only the operating points to its right satisfy the SLO.

For a NeoCloud provider that sells tokens through serverless endpoints, the height of the frontier at the SLO line is the per-GPU throughput that the provider can sell under that SLO, and because its GPUs and power are fixed costs, this height sets both its revenue capacity and its cost per token. The whole curve matters rather than a single point, because a provider serves both batch jobs (left) and latency-critical coding agents (right). Serving optimization therefore means pushing the frontier up and to the right, so that the same hardware delivers more tokens at every SLO.

{\bf Optimization space.}
An LLM serving stack consists of the model, a serving engine such as vLLM~\cite{vllm} or SGLang~\cite{sglang} that schedules requests and manages GPU memory, the GPU kernels that execute the model's layers, and the GPUs and interconnect beneath them. The engine uses continuous batching~\cite{orca}, which re-forms the batch at every iteration, so the batch size and its sequence lengths change from one iteration to the next, and with them the tensor shapes on which each GPU kernel operates (its \emph{input shapes}). The time of an iteration is the sum of GPU kernel execution, inter-GPU communication, and host-side (CPU) work such as request scheduling and kernel launches. Where a deployed serving stack, which we call a \emph{serving system}, lands in \fig{fig:pareto_frontier} therefore depends on the model, the hardware, the workload, and two kinds of decisions that the provider controls.

\emph{Configuration tuning} chooses values for the options, or \emph{knobs}, that the serving stack already exposes, without changing its code. The main knobs are the parallelism strategy, which determines how the model and its requests are distributed across GPUs through tensor (TP), pipeline (PP), data (DP), and expert (EP) parallelism, as well as decode context parallelism~\cite{DCPpaper}; prefill--decode disaggregation, which runs the two phases on separate instances~\cite{distserve,splitwise}; speculative decoding and the number of tokens it drafts~\cite{eagle,deepseekv3}; key--value (KV) cache policies such as prefix caching and offloading~\cite{sglang,lmcache}; the batching limits, i.e., the maximum numbers of sequences and tokens per iteration, and the batch sizes for which CUDA graphs are captured; and the kernel library, or \emph{backend}, that implements the attention and mixture-of-experts (MoE) layers.
\emph{Code optimization} changes the serving stack itself, for example by rewriting the engine's request scheduler or KV-cache manager, writing model-specific GPU kernels, fusing adjacent kernels, or trimming host-side overhead.

These two kinds of decisions depend on each other, because parallelism and batching settings determine kernel input shapes and communication patterns, and thus how much a faster kernel is worth; a new kernel or scheduling policy, conversely, shifts the bottleneck and can change which configuration is best. Consequently, neither can be optimized in isolation, and every change must be judged by its end-to-end effect, that is, by how far it moves the frontier. The best combination also depends on the model, the hardware, the workload, and the SLO, so a change in any of them reopens the search.

\subsection{Agent-Driven Serving Stack Optimization}
\label{subsect:agent_optimization}

Searching this optimization space manually is slow, and the search must be repeated every time the space is reopened. A growing number of techniques therefore delegate serving stack optimization to AI coding agents, i.e., LLMs that iteratively edit code, run tools, and act on the results~\cite{react,sweagent}, because an agent can carry out every step of the search by itself, namely proposing a change, implementing it, measuring its effect, and deciding what to try next.

{\bf Kernel optimization.}\label{subsect:agent_kernel}
GPU kernels were the first target, because the optimization task is well posed for a kernel, in that a reference implementation defines correctness and running the kernel in isolation measures its speed. Many benchmark suites and agent systems now follow this approach~\cite{kernelbench,stark,astra,accelopt} (\sect{subsect:related}), and agent-written kernels are deployed in hyperscale production~\cite{liao2026kernelevolve}. A faster kernel does not, however, guarantee faster serving, for two reasons. First, kernel microbenchmarks use a limited set of synthetic inputs, which reproduce neither the input shapes that vary with batch composition and context length in serving nor workload-dependent behavior such as the uneven loading of experts in an MoE model~\cite{lina,fastkernels}. Second, kernel execution is only part of each iteration, so, by Amdahl's law, the time spent in communication and host-side work caps the end-to-end benefit of any kernel speedup. Indeed, kernels that ran $1.6$--$6.6\times$ faster in isolation yielded at most a $1.25\times$ end-to-end speedup~\cite{fastkernels}.

{\bf Serving stack optimization.}\label{subsect:agent_serving}
More recent work therefore widens the agents' scope from individual GPU kernels to the entire serving stack, where coding agents plan, implement, and benchmark both configuration and code changes in a loop and judge each change by its end-to-end effect~\cite{kernelopt,serverbench2026}.

\begin{figure}[t]
  \centering
  \includegraphics[width=0.99\columnwidth]{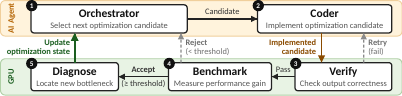}
    \vspace{0.3em}

  \captionsetup{font=small}
  \caption{Representative workflow of agent-driven serving-stack optimization, which we refer to as the baseline loop. Stages \circled{3}--\circled{5} run on the target GPUs.}
  \label{fig:agent_loop}
  \vspace{1em}
\end{figure}

\fig{fig:agent_loop} shows a representative workflow of such a loop.
A main agent, the \emph{orchestrator}, selects the next optimization \emph{candidate}, i.e., a single change to a configuration knob or to the code of the serving stack that is accepted or rejected as a whole (\circled{1}). A subagent, the \emph{coder}, which the orchestrator spawns for each candidate that changes code, then implements the candidate (\circled{2}), whereas a candidate that changes only a knob needs no implementation and is applied by the orchestrator itself.
The implementation is first tested on GPUs for output correctness (\circled{3});
a failing candidate is returned to the coder with the error.
A passing candidate is then \emph{benchmarked} end to end on GPUs (\circled{4}), and
it is rejected if its performance gain falls below a threshold and accepted otherwise.
After each acceptance, the updated serving system is \emph{diagnosed}, i.e., its use of time and resources is analyzed to locate its new bottleneck (\circled{5}),
and the orchestrator updates its optimization state with the diagnosis results to select the next candidate.
We refer to this workflow, in which every candidate is benchmarked and every accepted candidate is diagnosed on GPUs, as the \emph{baseline loop}. Existing systems focus on the serving performance that this loop reaches, i.e., the per-GPU throughput and interactivity of the optimized serving stack. The time that the loop takes to reach it, i.e., its time-to-efficiency, has received far less attention, even though three of its five stages must wait for the target GPUs. The rest of this section characterizes this time from the standpoint of a service provider that operates serverless LLM endpoints and shows that it is limited by two steps, namely benchmarking every candidate on GPUs (\sect{subsect:challenge2}) and diagnosing the serving system on GPUs before selecting the next candidates (\sect{subsect:challenge_coder_orchestration}).

%% file: tex/motivation.tex
\subsection{Time Spent on GPU Benchmarking}
\label{subsect:challenge2}
\begin{figure}[t]
  \centering
  \includegraphics[width=\columnwidth]{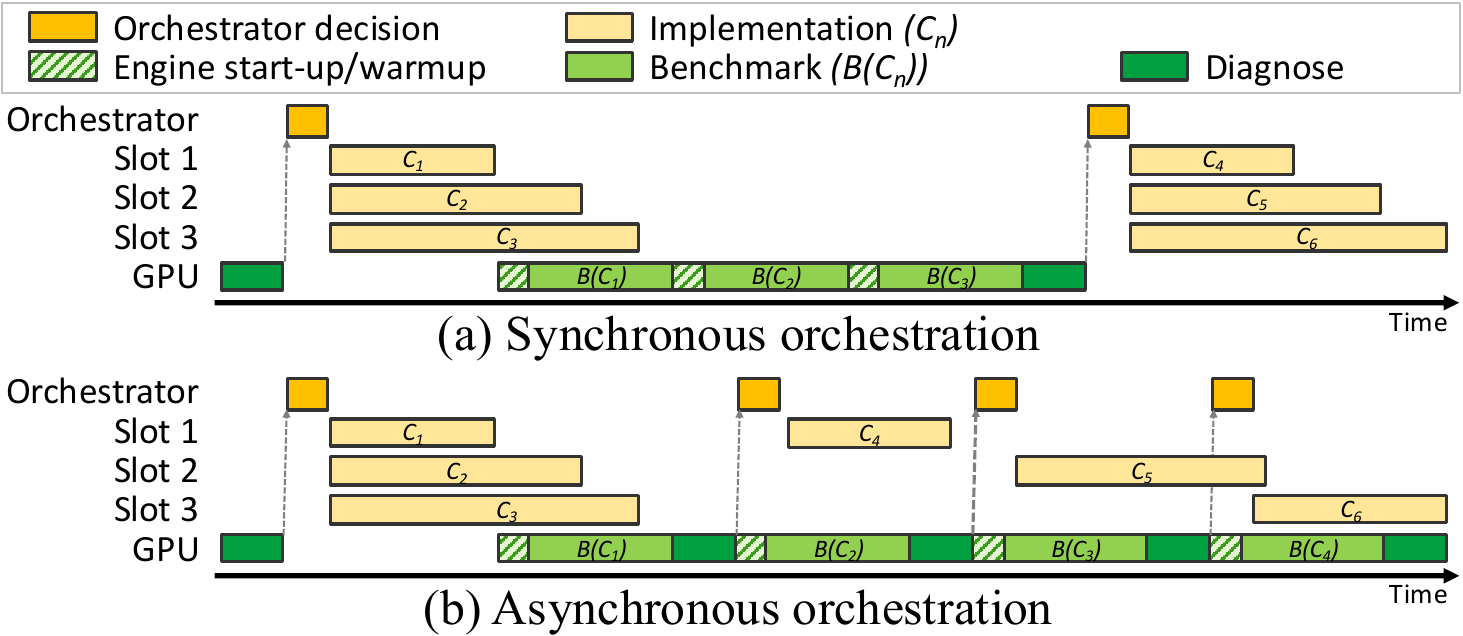}
 %   \vspace{0.3em}
  \captionsetup{font=small}
  \caption{Execution timelines of the baseline loop under (a) synchronous and (b) asynchronous orchestration.
  $C_n$ denotes the $n$-th candidate, and $B(C_n)$ its benchmarking on GPUs.}
  \label{fig:baseline_timeline}
    \vspace{1em}

\end{figure}

Agent-driven optimization loops make it fast to propose and implement candidates, since an AI agent can diagnose a performance bottleneck, propose a new kernel or configuration change, and implement it with little human effort, and multiple agents can do so in parallel. Finding out whether a candidate actually improves the serving system, however, has not become faster. This step is \emph{benchmarking}, in which the candidate's end-to-end performance, i.e., its throughput and interactivity across concurrency levels (\sect{subsect:pareto_frontier}), is measured under a workload that represents the target use case (e.g., AgentX~\cite{agentx} when the target is agentic AI) in order to decide whether to accept it (stage~\circled{4} in \fig{fig:agent_loop}). To be benchmarked, every candidate must be compiled, loaded into the serving engine, and run end to end, and these steps require GPUs. As a result, parallel coders can produce candidates faster than they can be benchmarked (\fig{fig:baseline_timeline}), and time-to-efficiency becomes limited not by the agents' ability to write code but by the time spent waiting for GPU benchmarking. We refer to this limit as the \emph{GPU benchmarking bottleneck}. It stems from three properties of GPU-based benchmarking, namely (i) broad benchmarking scope, (ii) long fixed time per benchmark sweep, and (iii) limited benchmarking capacity.

\begin{figure}[t]
  \centering
  \includegraphics[width=1.0\columnwidth]{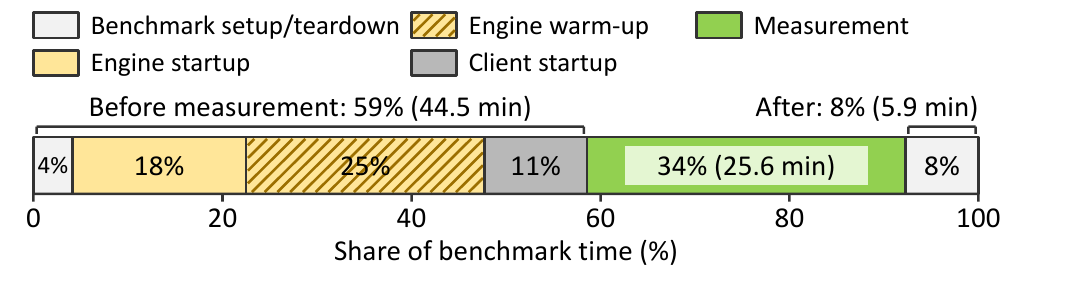}
 %   \vspace{0.3em}
  \captionsetup{font=small}
  \caption{Time breakdown of a single GPU benchmark sweep of GLM-5.3-Flash on an 8$\times$H100 server.}
  \label{fig:bench_overhead}
    \vspace{1em}

\end{figure}

\textbf{Broad benchmarking scope.}
The time spent on system-level benchmarking grows with the product of the number of candidates and the number of concurrency levels at which each candidate must be measured, not with the number of candidates alone. A candidate that helps at one concurrency level can hurt at another, because the concurrency level determines the batch size and thus which part of the serving system limits performance. A kernel tuned for the small batches of low concurrency, for example, can run slower than the original kernel on the large batches of high concurrency. Accepting a candidate therefore requires a sweep over the concurrency levels that the provider expects to serve, which traces the candidate's throughput--interactivity curve (\sect{subsect:pareto_frontier}), and each level is a separate benchmark run that takes several minutes at steady state. In our GLM-5.3-Flash experiments, benchmarking a single candidate across five concurrency levels takes about 1.3 hours on average on an 8$\times$H100 server (\fig{fig:bench_overhead}).

\textbf{Long fixed time per benchmark sweep.}
Each benchmark sweep spends a long, fixed time on setup before measurement begins. Most settings of the serving stack, such as the parallelism degree, speculative decoding settings, and kernel implementations, are fixed at engine startup, so changing any of them requires restarting the engine.
On an 8$\times$H100 server, a restart involves launching a worker process on each GPU and initializing inter-GPU communication, loading model weights, measuring memory usage to size the KV cache, and capturing CUDA graphs, and each concurrency level then requires its own engine warm-up before measurements stabilize, as well as the startup of a new load-generating client. \fig{fig:bench_overhead} shows the resulting time breakdown for GLM-5.3-Flash, averaged over 14 benchmark sweeps that each measure five concurrency levels. Preparation for measurement, including engine startup, engine warm-up, and client startup, takes 44.5 minutes (59\%), whereas the measurement itself takes only 25.6 minutes (34\%). Over an optimization run, this time adds up, as a 24-hour run of the baseline loop shows. In this run, engine startup and engine warm-up occupied the server for 8.0 hours, compared with 6.4 hours of measurement. Including the remaining setup and teardown steps, benchmark sweeps occupied the server for 18.8 hours, or nearly 80\% of the 24-hour optimization run.

\textbf{Limited benchmarking capacity.}
A benchmark sweep occupies the entire target server, and a provider can set aside only a few servers for optimization, because a server that runs benchmark sweeps cannot serve customer traffic. Benchmark sweeps therefore run one after another on a small, shared pool of GPUs, and the candidates that parallel coders produce queue for these GPUs. Much of this time is spent on candidates that are eventually discarded: in our experiments, of the 17 candidates sent to GPU benchmarking, 8 were rejected and 3 failed with engine startup or measurement errors, and these 11 discarded candidates occupied the server for 11.3 hours.

Together, these properties imply that rejecting a candidate costs nearly as much GPU time as accepting one, and that all candidates contend for the same few GPUs. \emph{Shortening time-to-efficiency therefore requires benchmarking candidates without occupying the target GPUs, so that GPU benchmarking is reserved for validating accepted candidates.}

\subsection{Time Spent Waiting for GPU Diagnosis}
\label{subsect:challenge_coder_orchestration}

Integrating an accepted candidate can shift the major performance bottlenecks of the serving system, which changes both which candidates are most promising and how valuable the coders' in-progress work is. Selecting subsequent candidates therefore requires \emph{diagnosing} the serving system again (stage~\circled{5} in \fig{fig:agent_loop}), i.e., analyzing where it spends its time (e.g., in GPU kernels or in host-side work) and how it uses its resources (e.g., the KV cache) to locate the new bottleneck. Whereas benchmarking decides the fate of a candidate that has already been implemented, diagnosis thus decides which candidates should be implemented next, and it is the second step for which the loop waits on GPUs. The orchestrator can start new candidates, and spawn coders for those that change code, in two ways, which we refer to as \emph{synchronous} and \emph{asynchronous} orchestration (\fig{fig:baseline_timeline}), and, as we show below, the dependency of candidate selection on diagnosis lengthens time-to-efficiency under both.

\begin{figure}[t]
  \centering
  \includegraphics[width=0.99\columnwidth]{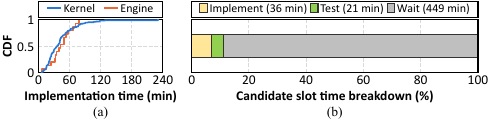}
        \vspace{0.3em}
  \captionsetup{font=small}
  \caption{Implementation time and candidate slot usage when optimizing GLM-5.3-Flash serving on 8$\times$H100. (a)~CDF of the time that a coder takes to implement a kernel or engine candidate, i.e., a change to GPU kernels or to the serving engine's code (\sect{subsect:proposed_overview}). (b)~Average time breakdown of a candidate slot in the baseline loop with synchronous orchestration.}
  \label{fig:coding_latency}
    \vspace{1em}

\end{figure}

\textbf{Inefficiencies of synchronous orchestration.}
The baseline loop uses synchronous orchestration, in which the optimization proceeds as a sequence of \emph{rounds} (\fig{fig:baseline_timeline}(a)). A round is one pass through the workflow of \fig{fig:agent_loop}, in which the orchestrator diagnoses the serving system, selects $N$ candidates based on the diagnosis results, and starts them, and each candidate is benchmarked on GPUs as soon as it has been implemented. At most $N$ candidates are thus in progress at the same time, and we refer to these $N$ positions as \emph{candidate slots}, each of which is freed as soon as its candidate has been accepted or rejected. Because a coder is spawned only for a candidate that changes code, at most $N$ coders run at the same time as well. A new round can begin only after every candidate of the previous round has been benchmarked, because the diagnosis with which it starts must reflect all of the candidates that the previous round accepted. We refer to this requirement as the \emph{synchronization barrier} for diagnosing the updated serving system, and, coupled with variation in implementation time across candidates, it leaves the candidate slots whose candidates finish early idle until the next round. Indeed, implementation times vary widely (\fig{fig:coding_latency}(a)), in that candidates that change GPU kernels (kernel candidates) take 6--231 minutes to implement, whereas candidates that change the serving engine (engine candidates) take 7--78 minutes. To quantify these inefficiencies, we run the baseline loop with synchronous orchestration for 24 hours, with $N=8$ candidate slots, targeting GLM-5.3-Flash on vLLM 0.32.0. As shown in \fig{fig:coding_latency}(b), a candidate slot spends on average only 36 minutes on implementation and 21 on testing, and it remains idle for the remaining 449 minutes, i.e., 89\% of the time.

\textbf{Limitations of asynchronous orchestration.}
One could relax the synchronization barrier through asynchronous orchestration, which does not divide the optimization into rounds and instead starts a new candidate as soon as a running candidate has been benchmarked, without waiting for the other coders (\fig{fig:baseline_timeline}(b)). The orchestrator cannot, however, propose the next candidate until it has located the bottleneck of the updated serving system, so this approach requires GPU-based diagnosis after every accepted candidate rather than once per round. Because each diagnosis occupies the same GPUs on which the other candidates are queued for benchmarking, it delays them further and aggravates the GPU benchmarking bottleneck (\sect{subsect:challenge2}). \emph{Shortening time-to-efficiency therefore also requires diagnosing the serving system without occupying the target GPUs, so that a new candidate can start as soon as a candidate slot is freed, without a synchronization barrier and without delaying GPU benchmarking.}

%% file: tex/proposed.tex
\section{\proposed{}: Agentic Serving Stack Optimizer for Accelerated Time-to-Efficiency}
\label{sect:proposed}

The two challenges identified in \sect{sect:motivation} have the same cause, which is that the baseline loop can benchmark a candidate and diagnose the serving system only on GPUs. \proposed{} alleviates both by co-designing coding agents and an LLM serving simulator, \simulator{}, in a closed loop. In one direction of this loop, the agents keep \simulator{} synchronized with the serving stack that they keep changing (\sect{subsect:a2sim}), and in the other direction, the predictions of \simulator{} take over most benchmarking (\sect{subsect:proposed_exploration} and \sect{subsect:proposed_gpu_validation}) and diagnosis (\sect{subsect:aync_coder_orchestration}) from the GPUs.

\subsection{Overview}
\label{subsect:proposed_overview}

\begin{figure}[t]
  \centering
  \includegraphics[width=\columnwidth]{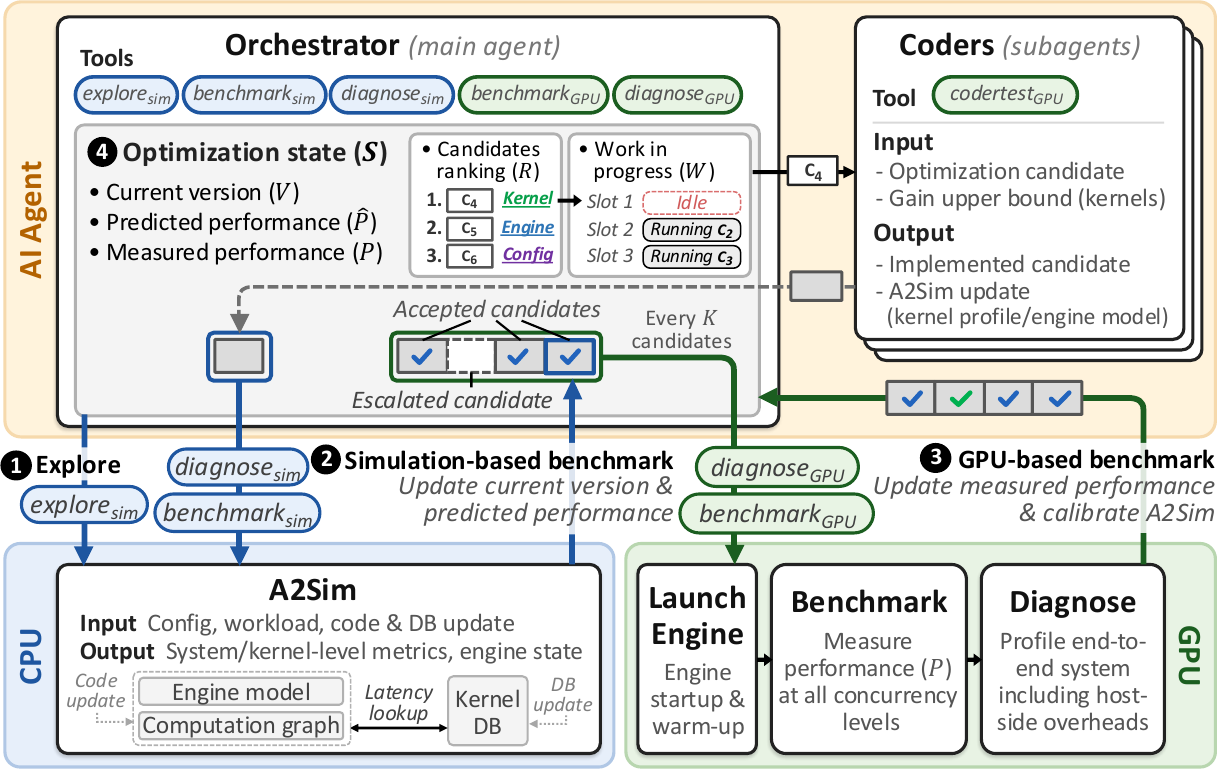}
%    \vspace{0.2em}
  \captionsetup{font=small}
  \caption{High-level overview of \proposed{}.}
  \label{fig:proposed_overview}
  \vspace{1em}
\end{figure}

\textbf{Agent architecture in \proposed.}
\fig{fig:proposed_overview} illustrates the overall structure and workflow of \proposed. As in the workflow of \sect{subsect:agent_serving}, \proposed employs a main agent, the \emph{orchestrator}, that manages the overall optimization process, and subagents, the \emph{coders}, each of which the orchestrator spawns to implement a single optimization candidate and which run in parallel, at most one per candidate slot (\sect{subsect:challenge_coder_orchestration}).
The orchestrator maintains an \emph{optimization state} $\mathcal{S} = (V, \hat{P}, P, R, W)$, which is its record of the optimization so far (\fig{fig:proposed_overview}). Here, $V$ is the current version of the serving system (its configuration and the candidates accepted so far), $\hat{P}$ is the performance that \simulator predicts for $V$, and $P$ is the performance most recently measured on GPUs, each of which comprises the benchmarking results and the diagnosis results. The other two components are the ranking $R$ of the remaining candidates and the work in progress $W$, i.e., the candidate that each running coder is currently implementing.
Each coder implements its candidate and verifies the correctness of the implementation on GPUs, which it also uses to profile the kernels that it modifies. A coder does not, however, benchmark end-to-end serving performance, and it returns the candidate to the orchestrator together with the update that \simulator{} needs to predict the serving performance with the candidate applied, i.e., new kernel latency profiles or a matching change to the code of \simulator{} (\sect{subsect:a2sim}).
The agents access \simulator{} simulations and GPU benchmarking through command-line interface (CLI) tools that they invoke from a bash shell (\fig{fig:proposed_overview}). The orchestrator explores the configuration space with \exploresim. It benchmarks and diagnoses the serving system with \benchmarksim and \diagnosesim in \simulator{} and with \benchmarkgpu and \diagnosegpu on GPUs. Each coder runs lightweight tests of its implementation on GPUs with \codertestgpu.
The agents also follow optimization procedures that are provided as skills, each of which is defined in a \texttt{SKILL.md} file~\cite{agentskills}.

\textbf{Optimization candidates.}
Candidates fall into three classes, which correspond to serving configuration tuning and to the two parts of the serving stack that code optimization changes (\sect{subsect:serving_optimization}). A \emph{configuration candidate} changes the values of serving configuration knobs (e.g., the batching limits or the size of the KV cache pool), a \emph{kernel candidate} changes GPU kernels (e.g., by specializing a kernel for the target model or by fusing adjacent kernels), and an \emph{engine candidate} changes the code of the serving engine (e.g., how it forms batches or preempts requests). Kernel and engine candidates change code and are therefore implemented by coders, whereas the orchestrator benchmarks a configuration candidate in \simulator{} directly without involving a coder.
Configuration candidates remain necessary after the starting serving configuration has been selected (\sect{subsect:proposed_exploration}), since an accepted code change shifts the bottleneck and can change which configuration is best (\sect{subsect:serving_optimization}).

\begin{algorithm}[t]
\caption{\proposed}
\label{alg:a2optimizer}\label{alg:batched_gpu}
\small
\begin{algorithmic}[1]
\Require initial optimization state $\mathcal{S}_0=(V_0,\hat{P}_0,P_0,R_0,W_0)$; candidate slots $N$, GPU benchmarking batch size $K$, acceptance threshold $\alpha$, time budget $T$
\Ensure optimized serving system $V$
\State $\mathcal{S} \gets \mathcal{S}_0$; $A \gets \emptyset$; $E \gets \emptyset$ \Comment{accepted and escalated candidates}
\State $V \gets \exploresim(V_0)$\label{algl:explore} \Comment{\S\ref{subsect:proposed_exploration}}
\State $\hat{P} \gets \benchmarksim(V)$; $\hat{P} \gets \diagnosesim(V)$
\State rank $R$; start candidates from $R$ in the $N$ slots; update $W$\label{algl:init}
\While{elapsed time $< T$}
  \State $e \gets$ wait until a candidate or GPU benchmarking completes
  \If{$e = \textsc{CandidateCompleted}$}\label{algl:next}
    \State $c \gets$ the completed candidate; $W \gets W \setminus \{c\}$
    \If{$c$ is within \simulator's modeling scope}
      \State $\hat{P}_c \gets \benchmarksim(V + c)$\label{algl:benchsim}
      \If{$\hat{P}_c$ improves on $\hat{P}$ by more than $\alpha$}\label{algl:criterion}
        \State $V \gets V + c$; $\hat{P} \gets \hat{P}_c$; $A \gets A \cup \{c\}$\label{algl:accept} \Comment{accept}
        \State $\hat{P} \gets \diagnosesim(V)$\label{algl:diagsim}
      \Else
        \State discard $c$\label{algl:reject} \Comment{reject}
      \EndIf
    \Else
      \State $E \gets E \cup \{c\}$\label{algl:escalate} \Comment{escalate}
    \EndIf
  \ElsIf{$e = \textsc{GPUBenchmarkCompleted}$}\label{algl:gpuevent}
    \State $\hat{P} \gets \benchmarksim(V)$\label{algl:repredict} \Comment{calibrated \simulator}
  \EndIf
  \If{$|A| + |E| \geq K$ and the GPUs are idle}\label{algl:batch}
    \State $P \gets$ \Call{GPUBenchmark}{$E$}; $A \gets \emptyset$; $E \gets \emptyset$\label{algl:endbatch}
  \EndIf
  \State re-rank $R$ from $\hat{P}$ and $P$; coordinate ongoing tasks in $W$\label{algl:rerank}
  \State start candidates from $R$ in the free slots; update $W$\label{algl:start}
\EndWhile
\State \Return $V$
\Statex
\Function{GPUBenchmark}{$E$} \Comment{in the background, \S\ref{subsect:proposed_gpu_validation}}
  \State $P \gets \benchmarkgpu(V)$; calibrate \simulator\label{algl:gpustep1} \Comment{step 1}
  \ForAll{$c \in E$}\label{algl:gpustep2} \Comment{step 2}
    \State $P_c \gets \benchmarkgpu(V + c)$
    \If{$P_c$ improves on $P$ by more than $\alpha$}
      \State $V \gets V + c$; $P \gets P_c$\label{algl:gpustep2end}
    \EndIf
  \EndFor
  \State $P \gets \diagnosegpu(V)$; \Return $P$\label{algl:gpudiag}
\EndFunction
\end{algorithmic}
%\vspace{2em}
\end{algorithm}

\textbf{Overall workflow.}
\algo{alg:a2optimizer} summarizes the workflow of \proposed{}, which first explores the serving configuration space in \simulator{} to select a starting configuration (\circled{1} in \fig{fig:proposed_overview}; line~\ref{algl:explore}) and selects the first candidates based on the diagnosis of that configuration in \simulator{} (\sect{subsect:proposed_exploration}). Whenever a candidate is ready to be benchmarked (line~\ref{algl:next}), the orchestrator gives it one of three verdicts (\circled{2}). A candidate that \simulator{} can model is benchmarked in \simulator{} and, according to its predicted performance, is either \emph{accepted}, in which case it is integrated into $V$ and the updated $V$ is diagnosed in \simulator{}, or \emph{rejected} (lines~\ref{algl:benchsim}--\ref{algl:reject}). A candidate that \simulator{} cannot model is instead \emph{escalated}, which means that \simulator{} does not judge it at all and the decision on the candidate is handed over to GPU benchmarking (line~\ref{algl:escalate}). Once $K$ candidates have been accepted in \simulator{} or escalated, \proposed{} benchmarks them on the target GPUs in the background and then diagnoses the resulting $V$ there (\circled{3}; line~\ref{algl:batch}; \sect{subsect:proposed_gpu_validation}). The measured performance replaces $P$ and is used to calibrate \simulator{}, and the GPU diagnosis exposes the host-side overhead that \simulator{} does not model, which leads to candidates that reduce it (\sect{subsect:aync_coder_orchestration}). After every verdict, the orchestrator re-ranks the remaining candidates and starts new candidates in the freed candidate slots, without waiting for the other coders to complete their tasks (\circled{4}; lines~\ref{algl:rerank}--\ref{algl:start}; \sect{subsect:aync_coder_orchestration}). Steps \circled{2}--\circled{4} then repeat, and together they form the closed loop, in which \simulator's predictions steer the agents (\circled{2} and \circled{4}), while the agents' kernel latency profiles, their changes to the code of \simulator{}, and the GPU measurements (\circled{3}) keep \simulator accurate.

\subsection{\simulator{} and Its Co-Evolution with the Serving Stack}
\label{subsect:a2sim}

\begin{figure}[t]
  \centering
  \includegraphics[width=\columnwidth]{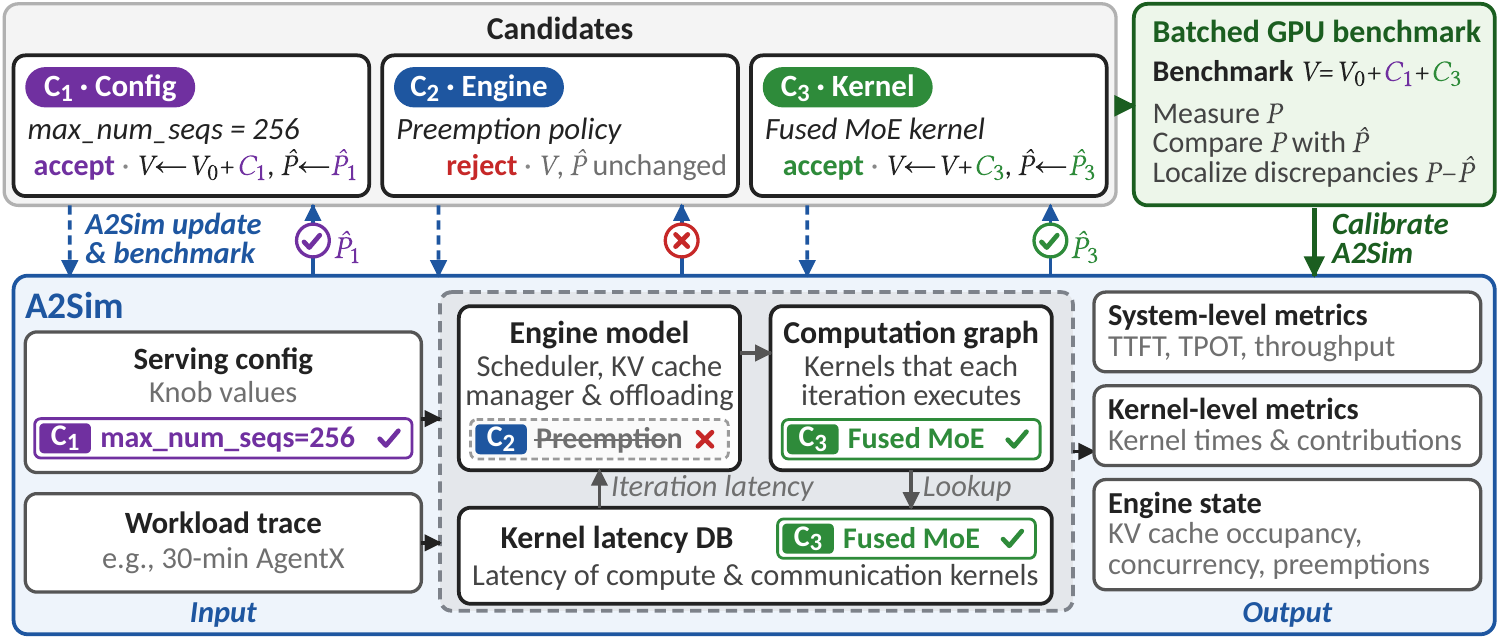}

  \captionsetup{font=small}
  \caption{\simulator{} and its co-evolution with the serving stack.
Each implemented candidate updates the \simulator{} component that models it and is benchmarked in \simulator{} on top of the current $V$, in the order $C_1$, $C_2$, $C_3$.
Accepted candidates ($C_1$, $C_3$) update $V$ and $\hat{P}$, while rejected ones ($C_2$) do not, so $C_1$ is applied to the initial serving system $V_0$ and $C_3$ to $V=V_0+C_1$.
Periodically, $V$ (here $V_0+C_1+C_3$) is benchmarked on GPUs to calibrate \simulator{}.}
  \label{fig:a2sim_coevolution}
  \vspace{1em}
\end{figure}

\textbf{Simulator design.}
\simulator, on which every step of the workflow above relies, employs the design philosophy of existing profile-driven, event-driven LLM serving~\cite{vserve2026, vidur,dynosim2026,llmservingsim,llmservingsim2} and training~\cite{vtrain,aiperf2026} simulators. It predicts serving system performance by modeling the behavior of the serving software stack and the performance of the underlying hardware, without executing the serving system on GPUs.
Specifically, each GPU computation and communication kernel is profiled \emph{offline} on the target GPUs across the batch sizes and sequence lengths that determine its input shapes, and the measured latencies are stored in a kernel latency database. The simulator then replays a workload trace through a model of the serving engine, and for every iteration that the modeled engine schedules, it composes the iteration latency from the database entries of the kernels that the iteration executes and advances its clock by that latency. This model of the serving engine, which we refer to as the \emph{engine model}, implements the same rules as the scheduler, the KV cache manager, and the KV cache offloading connector of the serving engine (vLLM~\cite{vllm} in our implementation), so that it makes the same decisions as the serving engine, e.g., which requests it batches, which it preempts, and how it allocates KV cache blocks. The kernels that each iteration executes are in turn given by \simulator{}'s \emph{computation graph} of the LLM.
Given a serving configuration, a workload trace, and the kernel latency database, \simulator{} produces three outputs (\fig{fig:a2sim_coevolution}, bottom). The first two are system-level metrics (e.g., per-request TTFT and TPOT, and token throughput) and a kernel-level latency breakdown (i.e., individual kernel execution times and their contributions to end-to-end inference time). The third is the state of the serving engine over time, such as the occupancy of the KV cache, the number of requests that run concurrently, and the preemptions that occur, which we refer to as the \emph{engine state}.
The system-level metrics are what benchmarking measures, and the latency breakdown and the engine state are what diagnosis analyzes (\sect{sect:motivation}).
\tab{tab:a2sim_scope} summarizes the modeling scope of \simulator{} for each candidate class (\sect{subsect:proposed_overview}).

\begin{table}[t]
  \centering
  \captionsetup{font=small}
  \caption{Modeling scope of \simulator by optimization candidate class
  (\textcolor{green!55!black}{\ding{51}}: modeled; \textcolor{red!75!black}{\ding{55}}: not modeled).
  \emph{Config} candidates change knob values, \emph{kernel} candidates change GPU kernels, and \emph{engine} candidates change the code of the serving engine.}
  \label{tab:a2sim_scope}
  \footnotesize
  \setlength{\aboverulesep}{0.5pt}
  \setlength{\belowrulesep}{0.5pt}
  \begin{tabular}{@{}l>{\raggedright\hangindent=0.8em\hangafter=1\arraybackslash}p{0.66\columnwidth}c@{}}
    \toprule
    \textbf{Class} & \textbf{Aspect (examples)} & \\
    \midrule
    \makecell[tl]{Config} & \textbf{Parallelism}: TP, PP, DP, EP, and decode context parallelism & \textcolor{green!55!black}{\ding{51}} \\
    \addlinespace[1pt]
     & \textbf{Batching limits}: maximum batch size and number of batched tokens & \textcolor{green!55!black}{\ding{51}} \\
    \addlinespace[1pt]
     & \textbf{KV cache capacity}: KV cache pool size and offloading tiers (host memory, SSD, and CXL memory) & \textcolor{green!55!black}{\ding{51}} \\
    \addlinespace[1pt]
     & \textbf{Engine features}: chunked prefill, asynchronous scheduling, prefix caching, and speculative decoding (EAGLE, MTP, and draft models, with the number of draft tokens) & \textcolor{green!55!black}{\ding{51}} \\
    \addlinespace[1pt]
     & \textbf{Kernel backends}: the libraries that implement the attention and MoE layers, and the communication collectives & \textcolor{green!55!black}{\ding{51}} \\
    \midrule
    \makecell[tl]{Kernel} & \textbf{Kernel changes}: new kernels, kernel fusion, overlapped execution on separate CUDA streams & \textcolor{green!55!black}{\ding{51}} \\
    \midrule
    \makecell[tl]{Engine} & \textbf{Scheduler}: continuous batching, and batch formation and preemption policies & \textcolor{green!55!black}{\ding{51}} \\
    \addlinespace[1pt]
     & \textbf{KV cache management}: block allocation, eviction, and prefix cache reuse policies & \textcolor{green!55!black}{\ding{51}} \\
    \addlinespace[1pt]
     & \textbf{KV cache offloading}: when and which blocks to load and evict across tiers & \textcolor{green!55!black}{\ding{51}} \\
    \addlinespace[1pt]
     & \textbf{Host-side overhead}: kernel launch, host--GPU synchronization, request scheduling overhead, and KV cache offloading software overhead & \textcolor{red!75!black}{\ding{55}} \\
    \bottomrule
  \end{tabular}
  \vspace{1.5em}
\end{table}

Simulating a serving system is much faster than benchmarking it on GPUs, as \simulator replays a 30-minute AgentX workload trace~\cite{agentx} for a single LLM serving instance in approximately 4 minutes, versus at least 30 minutes of GPU execution plus serving engine startup overhead.
Each simulation runs on a single CPU core, so many configurations can be benchmarked in parallel across cores.

\textbf{Reflecting optimization candidates in \simulator{}.}
\simulator{} depends on the serving stack both through its data, i.e., the kernel latency database, and through its code, i.e., the computation graph and the engine model (\fig{fig:a2sim_coevolution}). Before a candidate can be benchmarked in simulation, its change must therefore be reflected in whichever of these parts it affects. A configuration candidate requires no change to \simulator{}, because \simulator{} takes the knob values as an input and its engine model and computation graph already reflect how each knob changes the decisions of the serving engine and the kernels that each iteration executes (\tab{tab:a2sim_scope}). A kernel candidate that only makes existing kernels faster changes the data of \simulator{} alone, in that the latency profiles that its coder has collected are added to the kernel latency database as new variants of the modified kernels, next to the original entries. The remaining candidates require changes to the code of \simulator{}, which the coders make themselves, and there are two such cases. First, when a kernel candidate changes which kernels an iteration executes, e.g., by fusing several kernels into one, these kernels are replaced with the fused kernel in the computation graph, and \simulator{} computes its latency from the newly profiled database entries. Second, an engine candidate changes the decisions of the serving engine, e.g., its scheduling or KV cache rules, and a separate coder implements the same rules in \simulator{}'s engine model.

Whether a candidate changes the data or the code of \simulator{}, the change is added as an option that can be enabled, while the existing behavior of \simulator{} is left intact, so that \simulator{} can simulate the serving system both with and without the candidate, which is what the acceptance decision of \sect{subsect:proposed_gpu_validation} compares. A code change to \simulator{} is used for benchmarking only after \proposed{} has verified that \simulator{} with the option enabled produces the same results as the modified serving engine. For an engine candidate, this verification is a \emph{differential test}, in which the modified scheduling code of the serving engine and \simulator{} replay the same workload on CPUs and the decisions that the two make are compared at every scheduling step, including the composition of each batch, the status of each request, the preemptions, and the allocation, reuse, and offloading of KV cache blocks. Because the differential test runs on CPUs, it adds no GPU time to the loop. If the candidate is accepted, its option is enabled for $V$, so that every later candidate is simulated on top of it, and if the candidate is rejected, the option remains disabled, leaving the predictions for $V$ unchanged (\fig{fig:a2sim_coevolution}, top).

\subsection{Serving Configuration Space Exploration}
\label{subsect:proposed_exploration}

Before any code is written, \proposed{} selects a starting serving configuration, i.e., a value for each configuration knob of the serving stack, because these values determine where the bottlenecks of the serving system lie and thus which code changes are worth making (\sect{subsect:serving_optimization}). Simulating a configuration in \simulator{} removes the GPU run that benchmarking it would otherwise require, but the number of configurations grows exponentially with the number of knobs. Ten knobs with four values each, for example, already yield over a million configurations, which at four minutes per simulated sweep amounts to several years of CPU time. \proposed{} therefore explores the configuration space in two stages, a per-knob screening that discards the knobs and values that do not help on their own, followed by a cross-knob search over the combinations of the values that remain.

Both stages judge a configuration in the same way, relative to a reference configuration. The orchestrator simulates the configuration in \simulator{} with \exploresim at predefined concurrency levels (\sect{subsect:pareto_frontier}) and compares its per-GPU throughput and interactivity with those of the reference at each level. The configuration is \emph{beneficial} if either metric improves by more than $\alpha$ at one or more levels and neither metric degrades by more than $\alpha$ at any level, where $\alpha$ is the threshold that \proposed{} uses for every acceptance decision (\sect{sect:methodology}).

\textbf{Per-knob screening.}
The first stage starts from the default configuration of the target model (e.g., its vLLM recipe~\cite{vllm_recipe}) and varies one knob at a time while the other knobs keep their default values, a method known as \emph{one-factor-at-a-time screening}~\cite{czitrom1999ofat}. It keeps each value that is beneficial relative to the default configuration and drops every knob for which no value is. For example, sweeping the number of draft tokens in speculative decoding over 1 (the default), 2, 4, and 8 may keep only 2 and 4, and sweeping the maximum batch size over 64 (the default), 128, 256, and 512 may keep only 128 and 256. Because each knob is varied on its own, this stage simulates as many configurations as there are knob values in total rather than their product.

\textbf{Cross-knob search.}
Values that help individually can hurt when combined, an interaction that one-factor-at-a-time screening cannot reveal~\cite{czitrom1999ofat}. The second stage therefore simulates every combination of the values that the remaining knobs kept, together with each knob's default value so that a knob can also stay at its default. In the example above, it simulates the nine combinations of 1, 2, or 4 draft tokens with a maximum batch size of 64, 128, or 256. Among the combinations that are beneficial relative to the default configuration, the orchestrator selects the one with the largest gain at the target concurrency level, i.e., the level at which the provider expects to operate the endpoint, where the gain is the sum of the relative improvements in per-GPU throughput and interactivity over the default configuration. Ties are broken by the same sum over the other concurrency levels, and the default configuration is kept if no combination is beneficial. It then selects the first candidates based on \simulator{}'s diagnosis of the selected configuration, without waiting for this configuration to be benchmarked and diagnosed on GPUs, so that code optimization begins as soon as the exploration ends.

\subsection{Candidate Benchmarking}
\label{subsect:proposed_gpu_validation}

The baseline loop benchmarks every candidate on GPUs, which makes GPU benchmarking its bottleneck (\sect{subsect:challenge2}). \proposed{} instead decides on each candidate in \simulator{} and uses GPUs only for batches of candidates, both to validate the accepted candidates together and to decide on the few candidates that \simulator{} cannot model. Implementing a candidate means writing its code, testing and profiling this code on GPUs (\sect{subsect:proposed_overview}), and preparing \simulator{} to simulate it (\sect{subsect:a2sim}), all without applying the change to $V$, which happens only if the candidate is accepted. Because none of these steps involves GPU benchmarking over the target workload, and because \simulator{} can simulate the candidate as soon as it is implemented, a candidate within \simulator{}'s modeling scope receives its verdict within minutes instead of waiting hours in a queue for a GPU benchmark sweep of its own.

\textbf{Per-candidate decisions in simulation.}
When a candidate is ready, i.e., implemented by a coder or, for a configuration candidate, chosen by the orchestrator, the orchestrator first checks whether \simulator{} can model it (\tab{tab:a2sim_scope}). If it can, the orchestrator runs a full benchmark sweep of $V$ with the candidate applied in \simulator{}, using \benchmarksim, because a change that improves the part of the serving system that it touches does not necessarily improve the system as a whole (\sect{subsect:agent_kernel}). Such a sweep takes 4 minutes on a CPU instead of 1.3 hours on the GPU server. The orchestrator compares the result with the performance of $V$ without the candidate at each concurrency level, using the criterion and threshold $\alpha$ of \sect{subsect:proposed_exploration}. It accepts the candidate and integrates it into $V$ if the candidate is beneficial, and it rejects the candidate if it improves neither metric by more than $\alpha$ at any level. If the candidate improves performance beyond $\alpha$ at some levels but degrades it beyond $\alpha$ at others, the orchestrator decides by weighing the target concurrency level most heavily. If \simulator{} cannot model the candidate, e.g., an engine candidate that reduces host-side overhead, the candidate is \emph{escalated}, i.e., the decision is handed over to GPU benchmarking. Each of the three verdicts (accepted, rejected, and escalated) frees the candidate's slot immediately, so that the loop never stalls while a candidate waits for GPUs.

\textbf{Batched GPU benchmarking.}
Accepted and escalated candidates are not benchmarked on GPUs one at a time. Instead, once $K$ of them have accumulated, where we refer to $K$ as the GPU benchmarking batch size, \proposed{} benchmarks the batch on GPUs in two steps and then diagnoses the resulting $V$ on GPUs, which all run in the background so that the loop does not wait for them.

In the first step (line~\ref{algl:gpustep1} in \algo{alg:a2optimizer}), \proposed{} uses \benchmarkgpu to benchmark $V$, i.e., the serving system with all accepted candidates applied, once on GPUs in the same way as the baseline loop benchmarks a single candidate (\sect{subsect:challenge2}). This step does not revisit the acceptance decisions, which \simulator{} has already made. It instead measures how far the performance $\hat{P}$ that \simulator{} predicts for $V$ deviates from the performance $P$ that $V$ achieves on GPUs, localizes the deviation to individual kernels, and calibrates \simulator{} accordingly, so that later candidates are simulated with the calibrated \simulator{}. Because one sweep covers all accepted candidates of the batch, the time for engine startup, warm-up, and the sweep is spent once per batch rather than once per candidate. A larger $K$ amortizes this time over more candidates but also lets more prediction error accumulate between calibrations, a trade-off that we evaluate in \sect{subsect:eval_sensitivity}.

In the second step (lines~\ref{algl:gpustep2}--\ref{algl:gpustep2end}), \proposed{} decides on the escalated candidates of the batch by stacking them on top of $V$ one at a time. It adds one escalated candidate, benchmarks the serving system on GPUs, and accepts or rejects the candidate by the same criterion and threshold $\alpha$, now applied to the GPU measurements, before it adds the next one, so that an accepted escalated candidate stays in $V$ while the next one is benchmarked. Each escalated candidate thus costs a GPU benchmark sweep of its own, which is why \proposed{} benchmarks individually only the candidates that \simulator{} cannot model, i.e., those that are escalated.

After both steps, \proposed{} diagnoses the resulting $V$ on GPUs with \diagnosegpu{} (line~\ref{algl:gpudiag}), which exposes the host-side overhead that \simulator{} does not model and leads to candidates that reduce it (\sect{subsect:aync_coder_orchestration}). GPUs thus remain the ground truth, in that every operating point that \proposed{} reports has been validated on hardware, as the definition of time-to-efficiency requires (\sect{sect:introduction}).

\subsection{Diagnosis and Asynchronous Orchestration}
\label{subsect:aync_coder_orchestration}

As \sect{subsect:challenge_coder_orchestration} showed, the orchestrator can select the next candidates only after the serving system has been diagnosed again, so the baseline loop either waits at a synchronization barrier (\fig{fig:baseline_timeline}(a)) or diagnoses the serving system on GPUs after every accepted candidate (\fig{fig:baseline_timeline}(b)), which aggravates the GPU benchmarking bottleneck. \proposed{} removes this dilemma by diagnosing the serving system in \simulator{}, which is cheap enough to run after every accepted candidate.

\begin{figure}[t]
  \centering
  \includegraphics[width=\columnwidth]{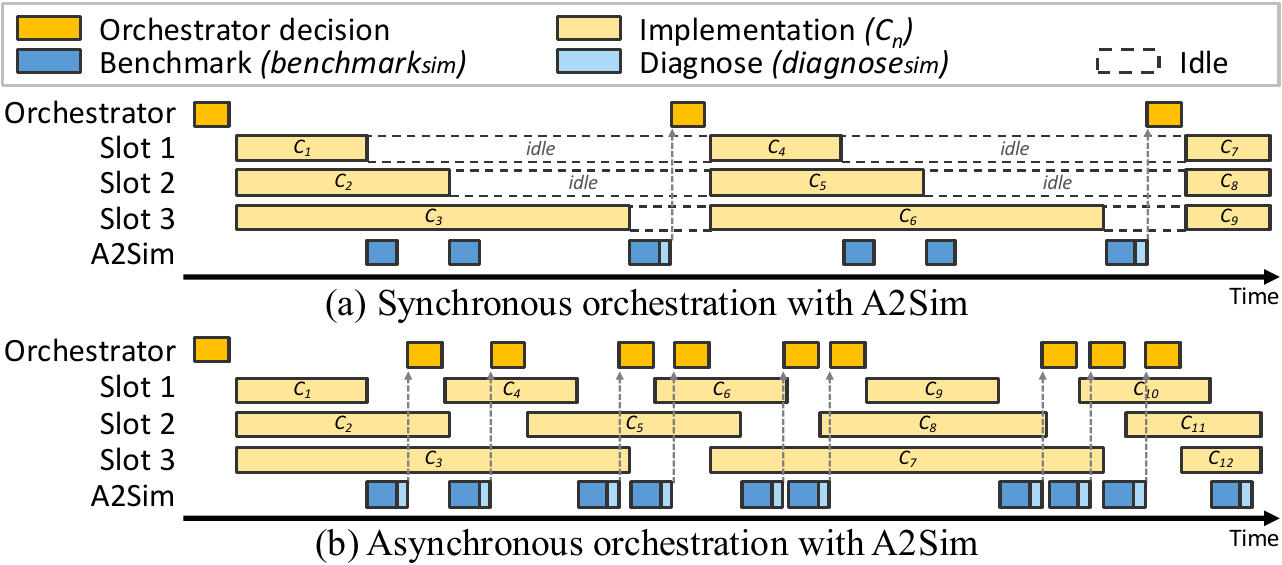}
  \captionsetup{font=small}
  \caption{Coder orchestration timelines with \simulator{}. (a)~Synchronous orchestration leaves the slots of early-finishing candidates idle until all candidates of the round have been implemented. (b)~\proposed{} enables asynchronous orchestration by using \simulator{} for benchmarking and diagnosis, allowing the orchestrator to update the optimization state $\mathcal{S}$ and start new candidates in the freed slots.}
  \label{fig:proposed_timeline}
    \vspace{1em}
\end{figure}

\begin{figure*}[t]
  \centering
  \includegraphics[width=0.85\textwidth]{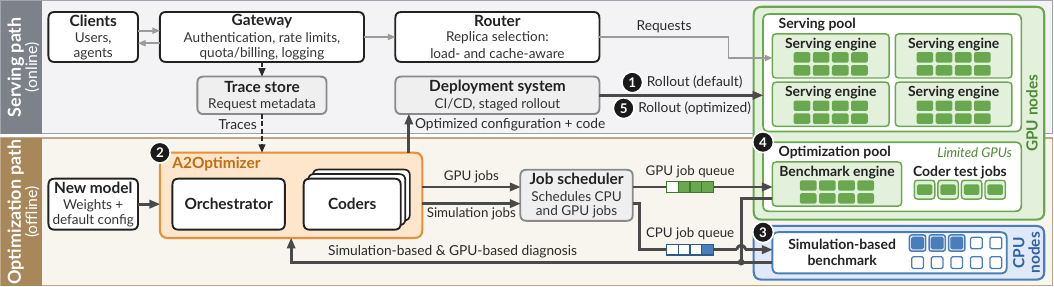}
  \vspace{0.7em}
  \caption{\proposed in a production serverless LLM endpoint service, with the online serving path (top) and the offline optimization path (bottom), on which \proposed turns a newly released model into an optimized serving stack (\circled{1}--\circled{5}).
  The optimization pool receives only a small share of the GPU nodes, most of which serve production traffic.}
  \label{fig:system_block}
  \vspace{1em}
\end{figure*}

\textbf{Simulation-based diagnosis.}
After every accepted candidate, the orchestrator diagnoses the updated $V$ in \simulator{} with \diagnosesim{} (line~\ref{algl:diagsim} in \algo{alg:a2optimizer}), which simulates $V$ on a CPU in minutes instead of profiling it on the target GPUs with \diagnosegpu{}. Because \simulator{} models the GPU kernels, the serving configuration, and the decisions of the serving engine (\tab{tab:a2sim_scope}), a single call to \diagnosesim{} exposes the bottlenecks behind all three classes of optimization candidates (\sect{subsect:proposed_overview}) through the two outputs of \simulator{} that diagnosis analyzes, i.e., the kernel-level latency breakdown and the engine state (\sect{subsect:a2sim}). First, the kernel-level latency breakdown, which resembles the one that PyTorch Profiler~\cite{torch_profiler} and NVIDIA Nsight Systems~\cite{nvidia_nsys} produce for a real serving system, shows which kernels dominate the iteration time and leads to kernel candidates. The orchestrator ranks these candidates in $R$ by an upper bound on the end-to-end improvement that optimizing each kernel can bring, i.e., the kernel's share of the iteration time multiplied by the headroom between its current latency and its roofline lower bound, which is the longer of the times that its ideal FLOPs and its minimum memory traffic would take at the GPU's peak compute rate and memory bandwidth. The orchestrator consults $R$ as a reference when it selects the next candidates to implement, and it gives the assigned coder the target kernel together with this upper bound, instead of a target speedup, to guide the implementation of the kernel candidate.

Second, the engine state shows how the serving engine uses its resources, e.g., how full the KV cache stays and how often requests are preempted, and leads to configuration and engine candidates. For example, a KV cache that remains full while requests are repeatedly preempted indicates that memory, not computation, limits throughput, which calls for a larger KV cache pool or a different preemption policy and which no kernel-level latency breakdown would reveal. Because configuration and engine candidates do not target an individual kernel, the upper bound by which kernel candidates are ranked cannot be computed for them. The orchestrator instead prioritizes configuration and engine candidates based on the engine state and ranks them in $R$ together with the kernel candidates (\fig{fig:proposed_overview}).

Simulation-based diagnosis cannot, however, expose what \simulator{} does not model, namely host-side overhead such as kernel launches, host--GPU synchronization, request scheduling, and calls to the KV cache offloading service. The orchestrator therefore complements the two outputs of \simulator{} with the GPU diagnosis at the end of each batched GPU benchmarking (\sect{subsect:proposed_gpu_validation}), which exposes this overhead and leads to candidates that reduce it. Because \simulator{} cannot model these candidates either, they are escalated and decided on GPUs one at a time.

\textbf{Asynchronous orchestration.}
Because diagnosis no longer requires GPUs, the orchestrator handles each event as soon as it occurs instead of proceeding in the rounds of synchronous orchestration (\fig{fig:proposed_timeline}). When a candidate receives its verdict (\sect{subsect:proposed_gpu_validation}), its slot is freed, and if the candidate was accepted, the orchestrator first diagnoses the updated $V$ in \simulator{}. It then re-ranks $R$ based on the updated optimization state $\mathcal{S}$ and starts the top-ranked candidate in the freed slot, spawning a coder if the candidate changes code (lines~\ref{algl:rerank}--\ref{algl:start}). When a batched GPU benchmarking completes, the orchestrator records the measured performance and the GPU diagnosis in $\mathcal{S}$, simulates $V$ again with the calibrated \simulator{} (line~\ref{algl:repredict}), and re-ranks $R$ in the same way. A freed slot is therefore refilled immediately, without waiting for other coders, GPU benchmarking, or GPU diagnosis, and because the target GPUs are used only for batched GPU benchmarking, asynchronous orchestration no longer aggravates the GPU benchmarking bottleneck as it does in the baseline loop.

\textbf{State-aware coder management.}
Because new candidates start while other coders are still working, a candidate accepted in the meantime can change the priority or the relevance of the tasks in progress. Beyond re-ranking $R$, the orchestrator therefore uses the work in progress $W$ in two ways. When the tasks of two coders overlap, e.g., when two coders optimize the same fused MoE kernel for decode and prefill, respectively, the orchestrator lets them share their modifications, which avoids redundant work and conflicting edits. When an accepted candidate substantially lowers the priority of a task in progress or makes it obsolete, e.g., when an accepted scheduling policy reduces the latency share of a kernel that another coder is optimizing, the orchestrator terminates the task or adjusts its objective.

\subsection{Putting Everything Together: Deploying \proposed{} in Serverless Endpoints}
\label{sect:case_study}\label{sect:casestudy}

We conclude this section by showing how \proposed{} is deployed in our production service, which offers serverless LLM endpoints for open-weight models. As \fig{fig:system_block} shows, the service consists of two paths, an online serving path and an offline optimization path.

\textbf{Online serving path.}
A \emph{gateway} records the metadata of every client request in a \emph{trace store}. To protect user privacy, the service follows a zero data retention (ZDR) policy and never stores user prompts or the responses to them. The trace store therefore holds only basic information about each request, such as its arrival time and its numbers of input and output tokens, and this information is permanently deleted after a fixed retention period. A \emph{router} forwards the request to one of the replicas of the requested model, each of which is a serving engine that occupies GPU nodes in the \emph{serving pool}. A \emph{deployment system} rolls out serving configurations and engine code to these replicas.

\textbf{Offline optimization path.}
When a new open-weight model is released, the deployment system first rolls it out with its default configuration (\circled{1}), so that the endpoint is available on the day of release. \proposed{} then optimizes the serving stack offline while this endpoint keeps serving (\circled{2}). Its inputs are the weights and default configuration of the model, together with the request traces in the trace store, which provide the workload under which candidates are benchmarked. The orchestrator and the coders submit their work to a \emph{job scheduler} as simulation jobs, which run \simulator{} in parallel on CPU nodes (\circled{3}), and as GPU jobs, which run in an \emph{optimization pool} (\circled{4}). This pool holds the correctness tests and kernel profiling of the coders and a benchmark engine that performs the GPU benchmarking of \sect{subsect:proposed_gpu_validation}. When the optimization ends, \proposed{} passes the optimized configuration and code to the deployment system, which rolls them out to the serving pool (\circled{5}). The optimization pool receives only a small share of GPU nodes, because every GPU assigned to it cannot serve requests. If every candidate had to be benchmarked on GPUs, they would all queue for the benchmark engine of this small pool, which is the GPU benchmarking bottleneck of \sect{subsect:challenge2} in its production form, whereas \proposed{} moves per-candidate benchmarking and diagnosis to CPU nodes, which are far less scarce than GPUs.

%% file: tex/methodology.tex
\section{Methodology}
\label{sect:methodology}

\textbf{Hardware and software.}
All GPU benchmarking and GPU diagnosis run on the target server, a single node with eight NVIDIA H100 80GB SXM GPUs and up to 1.7~TiB of CPU memory.
The coders share up to eight additional GPUs for the correctness tests and kernel profiling of their implementations (\sect{subsect:proposed_overview}).
\simulator runs as CPU-only jobs on separate CPU nodes with two Intel Xeon Platinum 8462Y+ processors (120~vCPUs) and 1.9~TiB of memory.
The serving stack to be optimized uses vLLM 0.32.0~\cite{vllm} as the serving engine and LMCache 0.5.6~\cite{lmcache} for KV cache offloading, and the orchestrator and the coders are Claude Code~\cite{claude_code} agents backed by Claude Opus 5.5~\cite{opus_5.5} with high reasoning effort.

\textbf{Models and workloads.}
\tab{tab:target_models} lists the four open-weight models that we optimize, all of which are served from their public checkpoints.
To cover both agentic and conversational serving, we optimize GLM-5.3-Flash, DeepSeek-V4.1-Flash, and Qwen3.8-Flash-Next under AgentX~\cite{agentx}, a multi-turn coding-agent workload with recorded request inter-arrival times, growing contexts, and substantial prefix reuse, and GPT-OSS-120B under ShareGPT, a general-purpose chat workload with short conversational requests.
For Qwen3.8-Flash-Next, we keep only the AgentX traces that fit within its 256K-token context window (220 of 393 traces).
Each model is measured at five concurrency levels, namely 1, 16, 32, 48, and 64 under AgentX and 1, 8, 32, 128, and 512 under ShareGPT, and a GPU benchmark sweep launches the serving engine once and measures each level for 5 minutes.

\begin{table}[t]
  \centering
  \scriptsize
  \caption{Target models and workloads.}
  \label{tab:target_models}
  \setlength{\tabcolsep}{4pt}
  \begin{tabular*}{\columnwidth}{@{\extracolsep{\fill}}lcccc@{}}
    \hline
     & \makecell{GLM-5.3-\\Flash~\cite{glm_5.3_flash}}
     & \makecell{DeepSeek-\\V4.1-Flash~\cite{deepseekv41flash}}
     & \makecell{Qwen3.8-\\Flash-Next~\cite{Qwen_3.8_flash_next}}
     & \makecell{GPT-OSS-\\120B~\cite{gpt_oss_120b}} \\
    \hline
    Params (tot./act.) & 321B/18B & 749B/8--16B & 176B/6B & 120B/5.1B \\
    Parallelism & TP8+EP & TP8+EP & TP8+EP & TP8 \\
    \hline
    Workload & AgentX & AgentX & AgentX & ShareGPT \\
    Concurrency & 1--64 & 1--64 & 1--64 & 1--512 \\
    \hline
  \end{tabular*}
    \vspace{1em}
\end{table}

\textbf{Optimization objective.}
Each run optimizes the throughput--interactivity frontier (\sect{subsect:pareto_frontier}) of the serving system, i.e., per-GPU output-token throughput against interactivity across these concurrency levels, where interactivity is computed from the p50 TPOT and every operating point must keep the p50 TTFT within 10~s.
Every acceptance decision follows the criterion of \sect{subsect:proposed_exploration} with a threshold of $\alpha=0.5\%$.
The target concurrency level (\sect{subsect:proposed_exploration}) is 32 for every model, and \sect{sect:evaluation} reports its results at this level.

\textbf{Compared systems.}
We compare \proposed{} with the baseline loop under the synchronous orchestration of \sect{subsect:challenge_coder_orchestration}, in which every candidate is benchmarked and the serving system is diagnosed once per round, both on the target server.
Both start from the same \emph{initial configuration}, i.e., the unmodified serving stack with the default configuration of the model's vLLM recipe~\cite{vllm_recipe}, and both use the same agents, acceptance threshold, target server, and 24-hour wall-clock budget.
Unless otherwise specified, every run uses $N=8$ candidate slots (\sect{subsect:challenge_coder_orchestration}), and \proposed{} uses a GPU benchmarking batch size of $K=8$ (\sect{subsect:proposed_gpu_validation}).
To isolate the effect of asynchronous orchestration, \sect{subsect:eval_ablation} additionally evaluates an asynchronous variant of the baseline loop and a synchronous variant of \proposed{}.

\textbf{Metrics.}
We report the progress of a run as its \emph{speedup}, i.e., the highest per-GPU throughput at the target concurrency level that the run has measured on the target server so far, relative to that of the initial configuration.
A speedup thus rises only when GPU benchmarking confirms an improvement, so a gain that \simulator{} merely predicts does not count.
We measure time-to-efficiency as the elapsed time at which a run first reaches a \emph{target speedup}, which is the highest speedup that the baseline loop reaches within the 24-hour budget unless stated otherwise.
We also report the \emph{GPU time per candidate}, i.e., the total time for which GPU benchmarking and GPU diagnosis occupy the target server divided by the number of benchmarked candidates.
Finally, because some candidates change the numerics of the serving system, e.g., kernels that reorder or fuse floating-point operations, we check model accuracy by evaluating the initial configuration and the final serving system of \proposed{} with five-shot prompting on all 1,319 test problems of GSM8K~\cite{cobbe2021gsm8k} (greedy decoding) and all 14,042 questions of MMLU~\cite{hendrycks2021mmlu} (log-likelihood scoring).

%% file: tex/evaluation.tex
\section{Evaluation}
\label{sect:evaluation}

\subsection{Time-to-Efficiency and Serving Performance}
\label{subsect:eval_main}

\begin{figure}[t]
    \centering
    \includegraphics[width=0.99\columnwidth]{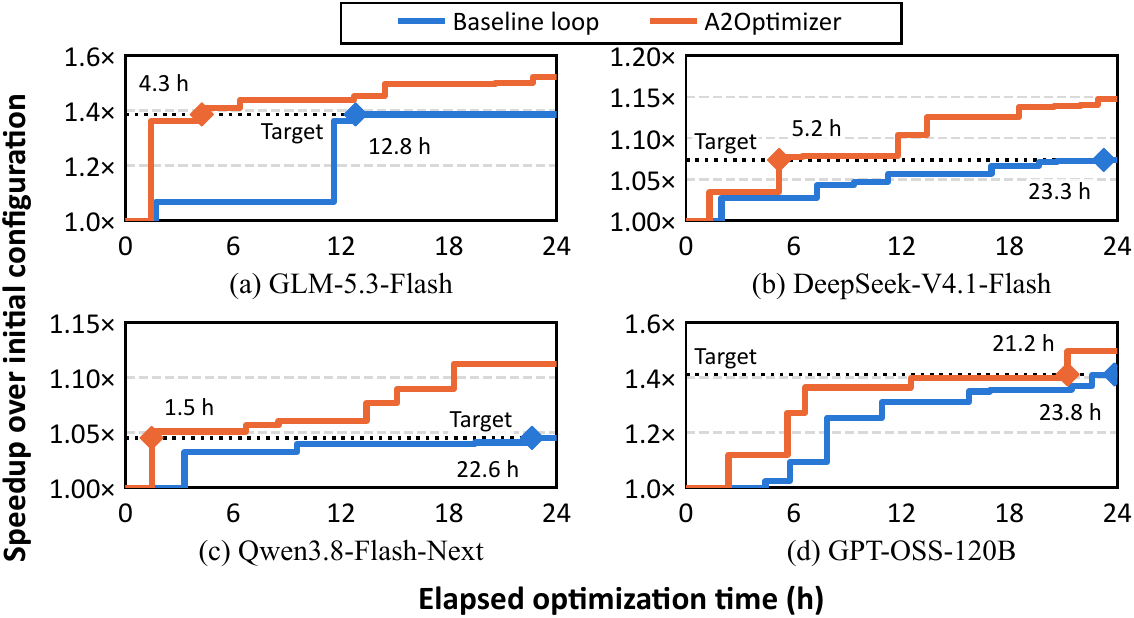}
    \captionsetup{font=small}
    \caption{Optimization progress of the baseline loop and \proposed{}, shown as the highest speedup in per-GPU throughput at the target concurrency level over the initial configuration that each system has measured on GPUs so far. Diamonds mark when each system first reaches the target speedup (black dotted lines).}
    \label{fig:time-to-efficiency}
      \vspace{1em}
\end{figure}

\fig{fig:time-to-efficiency} plots the speedup of the baseline loop and of \proposed{} over the 24-hour budget for the four models of \tab{tab:target_models}, together with the target speedup of each model (\sect{sect:methodology}).

\textbf{Time-to-efficiency.}
The baseline loop reaches the target speedup after 12.8--23.8 hours across the four models, whereas \proposed{} reaches it in 1.5--21.2 hours, which shortens time-to-efficiency by 1.1--15.1$\times$ (a geometric mean of 3.9$\times$).
For every model, \proposed{} holds a higher speedup than the baseline loop at every point in time after its first GPU benchmarking.
For GLM-5.3-Flash and DeepSeek-V4.1-Flash, \proposed{} reaches the target speedup in 4.3 and 5.2 hours, compared with 12.8 and 23.3 hours for the baseline loop.
The reduction is largest for Qwen3.8-Flash-Next, at 15.1$\times$, because the baseline loop barely improves on the initial configuration of this model, so the target speedup corresponds to an improvement of only 5\%, which the starting configuration that \proposed{} selects through its configuration space exploration (\sect{subsect:proposed_exploration}) already exceeds at 1.5 hours.
The reduction is smallest for GPT-OSS-120B, at 1.1$\times$, where \proposed{} comes within 1\% of the target speedup at 12.5 hours but does not exceed it until 21.2 hours, while the baseline loop reaches it at 23.8 hours.
These reductions stem from the GPU time that each candidate requires, since \proposed{} runs GPU benchmarking only for batches of accepted candidates and for the few escalated candidates, a difference that \sect{subsect:eval_ablation} analyzes in detail.

\begin{figure}[t]
    \centering
    \includegraphics[width=0.95\columnwidth]{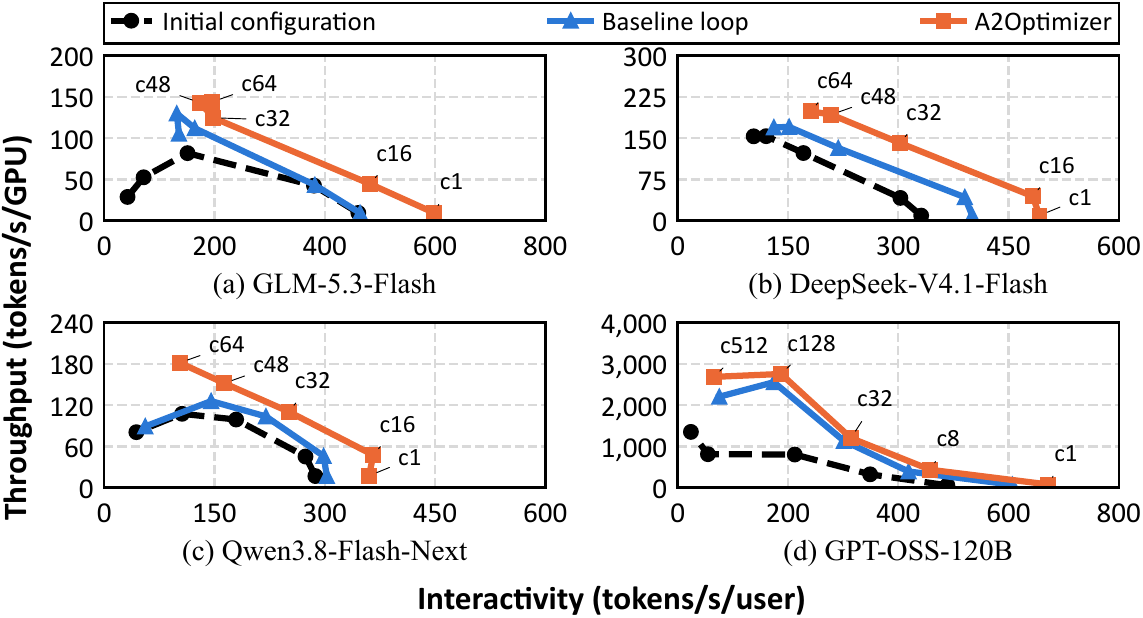}
    \captionsetup{font=small}
    \caption{Per-GPU throughput versus interactivity of the initial
    configuration and of the final serving systems of the baseline loop and
    \proposed{} after the 24-hour budget.
    Each point is one concurrency level, and the labels give the concurrency
    levels on the curve of \proposed{}, which the other two curves share.}
    \label{fig:serving-performance}
      \vspace{1em}

\end{figure}

\textbf{Serving performance.}
\fig{fig:serving-performance} compares the throughput--interactivity curves (\sect{subsect:pareto_frontier}) of the initial configuration with those of the final serving systems of the baseline loop and \proposed{}.
At the target concurrency level, \proposed{} improves per-GPU throughput by 1.12--1.52$\times$ and interactivity by 1.31--1.77$\times$ over the initial configuration, which is 1.06--1.11$\times$ and 1.04--1.38$\times$ higher than the baseline loop, respectively.
The gains are not limited to this level, since the curve of \proposed{} forms the throughput--interactivity frontier at every concurrency level.
The gains in per-GPU throughput over the initial configuration are largest at high concurrency levels, where the initial configuration saturates while \proposed{} continues to scale.
For GLM-5.3-Flash, for example, raising the concurrency level from 32 to 64 lowers the per-GPU throughput of the initial configuration from 81.9 to 28.8 tokens/s/GPU and its interactivity from 151.1 to 42.4 tokens/s/user, whereas \proposed{} raises per-GPU throughput from 124.8 to 144.1 tokens/s/GPU while keeping interactivity at about 197 tokens/s/user.
At concurrency levels of up to 16, \proposed{} instead improves mainly interactivity, by 1.3--1.6$\times$ over the initial configuration and by 1.1--1.3$\times$ over the baseline loop.

\textbf{Model accuracy.}
The final serving systems of \proposed{} preserve model accuracy, in that their GSM8K and MMLU scores differ from those of the initial configuration by at most 0.23 percentage points.

\subsection{Sensitivity Study}
\label{subsect:eval_sensitivity}

We next examine how sensitive \proposed{} is to its two main parameters, the number of candidate slots $N$ and the GPU benchmarking batch size $K$, using GLM-5.3-Flash.
Because the configuration space exploration (\sect{subsect:proposed_exploration}) does not depend on $N$ or $K$, both studies measure time from the end of the exploration and speedup relative to the starting configuration that it selects.
Each study sets its target speedup to the highest speedup that its worst-performing run reaches within the 24-hour budget.

\begin{figure}[t]
    \centering
    \includegraphics[width=0.99\columnwidth]{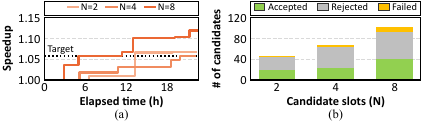}
    \captionsetup{font=small}
    \caption{Sensitivity to the number of candidate slots ($N$)
    for GLM-5.3-Flash: (a) time-to-efficiency, with time measured from the end of the configuration
    space exploration, and (b) numbers of accepted, rejected, and failed candidates.}
    \label{fig:sensitivity_n_candidates}
      \vspace{1em}

\end{figure}

\begin{figure}[t]
    \centering
    \includegraphics[width=0.99\columnwidth]{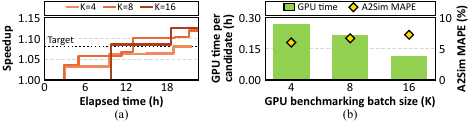}
    \captionsetup{font=small}
    \caption{
    Sensitivity to the GPU benchmarking batch size ($K$) for GLM-5.3-Flash: (a) time-to-efficiency, with time measured from the end of the configuration
    space exploration, and (b) GPU time per candidate
    (left axis) and MAPE of the per-GPU throughput that \simulator{} predicts before each batched
    GPU benchmarking calibrates it, relative to the GPU measurement (right axis).}
    \label{fig:sensitivity_bench_batch_size}
      \vspace{1em}

\end{figure}

\textbf{Number of candidate slots.}
We vary $N$ over 2, 4, and 8 with $K=8$, and as \fig{fig:sensitivity_n_candidates}(a) shows, $N=8$ yields the shortest time-to-efficiency, reaching the target speedup in 11.3 hours, whereas $N=2$ and $N=4$ reach it in 13.3 and 19.9 hours, respectively, which is why we use $N=8$ by default.
With a small $N$, few candidates are in progress at the same time, so a candidate that takes long to implement (\sect{subsect:challenge_coder_orchestration}) is not offset by others that finish in the meantime, and candidate implementation becomes the bottleneck of the loop.
\fig{fig:sensitivity_n_candidates}(b) breaks down the candidates that finish within the 24-hour budget into accepted, rejected, and failed candidates, where an escalated candidate counts as accepted or rejected according to its GPU benchmarking result, and a failed candidate is one that is never benchmarked, mostly because its implementation fails the coder's correctness test on GPUs (\codertestgpu).
Raising $N$ from 2 to 8 raises the number of accepted candidates from 19 to 41, but the number of accepted candidates per slot drops from 9.5 to 5.1, and the fraction of failed candidates grows from 4\% to 9\%.
A larger $N$ may thus shorten time-to-efficiency further, but it cannot be raised indefinitely, because every additional slot adds a coder that runs at the same time, with its cost and its use of the GPUs that the coders share for correctness tests, while yielding fewer accepted candidates per slot.

\textbf{GPU benchmarking batch size.}
We vary $K$ over 4, 8, and 16 with $N=8$, and as \fig{fig:sensitivity_bench_batch_size}(a) shows, a larger $K$ shortens time-to-efficiency, in that $K=16$ reaches the target speedup in 9.8 hours, compared with 13.0 and 18.9 hours for $K=8$ and $K=4$, respectively.
The reason is that a small $K$ runs batched GPU benchmarking more often and thus amortizes each GPU benchmark sweep over fewer candidates (\sect{subsect:proposed_gpu_validation}), so the GPU time per candidate rises from 0.11 hours at $K=16$ to 0.27 hours at $K=4$ (\fig{fig:sensitivity_bench_batch_size}(b)).
A larger $K$ also calibrates \simulator{} less often, yet the mean absolute percentage error (MAPE) of the per-GPU throughput that \simulator{} predicts before each calibration rises only slightly, from 6.0\% at $K=4$ to 7.3\% at $K=16$, because the coders keep feeding their kernel latency profiles back into \simulator{} between calibrations (\sect{subsect:a2sim}).
What a large $K$ does cost is GPU diagnosis, which runs once per batch and is the only way to expose what \simulator{} does not model, such as host-side overhead (\sect{subsect:aync_coder_orchestration}).
With $K=8$, for example, four of the six escalated candidates are accepted, most of which reduce host-side overhead and which together account for 42\% of the per-GPU throughput gain of the run, whereas with $K=16$ only three candidates are escalated and none is accepted.
A large $K$ can thus miss such optimizations, so we use $K=8$ by default to balance the cost of GPU benchmarking against this risk.

\subsection{Ablation Study}
\label{subsect:eval_ablation}

\begin{figure}[t]
    \centering
    \includegraphics[width=0.99\columnwidth]{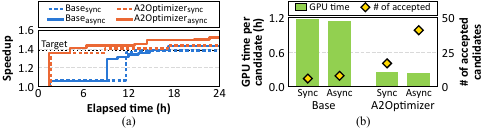}
    \captionsetup{font=small}
    \caption{Effect of asynchronous orchestration for GLM-5.3-Flash:
    (a) time-to-efficiency and (b) GPU time per candidate (left axis),
    with the number of accepted candidates (right axis).}
    \label{fig:ablation_orchestration}
      \vspace{1em}

\end{figure}

We finally isolate the effect of asynchronous orchestration by running both the baseline loop and \proposed{} with synchronous and with asynchronous orchestration, which we denote as \texttt{Base\textsubscript{sync}}, \texttt{Base\textsubscript{async}}, \texttt{\proposed{}\textsubscript{sync}}, and \texttt{\proposed{}\textsubscript{async}}, where \texttt{Base\textsubscript{sync}} and \texttt{\proposed{}\textsubscript{async}} are the two systems of \sect{subsect:eval_main}.
All four optimize GLM-5.3-Flash with $N=8$ candidate slots, and the two variants of \proposed{} use $K=8$.
\texttt{Base\textsubscript{async}} follows the asynchronous orchestration of \sect{subsect:challenge_coder_orchestration}, with a GPU diagnosis after every accepted candidate.
\texttt{\proposed{}\textsubscript{sync}} benchmarks every candidate in \simulator{} as \texttt{\proposed{}\textsubscript{async}} does, but it proceeds in rounds and therefore differs in three respects.
It waits for the starting configuration to be benchmarked and diagnosed on GPUs before it selects the first candidates, it diagnoses the serving system in \simulator{} once per round rather than after every accepted candidate, and it starts batched GPU benchmarking only at a round boundary, once at least $K$ candidates have been accepted or escalated.
As in \sect{subsect:eval_main}, the target speedup is the highest speedup that \texttt{Base\textsubscript{sync}} reaches.

\fig{fig:ablation_orchestration}(a) shows that asynchronous orchestration helps \proposed{} but not the baseline loop.
\texttt{\proposed{}\textsubscript{async}} reaches the target speedup 2.1$\times$ earlier than \texttt{\proposed{}\textsubscript{sync}}, whereas \texttt{Base\textsubscript{async}} reaches it after 14.8 hours, later than the 12.8 hours of \texttt{Base\textsubscript{sync}}.
\fig{fig:ablation_orchestration}(b) explains this difference through the GPU time per candidate and the number of candidates that each variant accepts within the 24-hour budget.
Both variants of the baseline loop spend about 1.2 hours of GPU time per candidate, so they remain limited by the GPU benchmarking bottleneck (\sect{subsect:challenge2}), and asynchronous orchestration raises the number of accepted candidates only from 6 to 8.
The GPU time per candidate of \proposed{}, by contrast, is only 0.23--0.26 hours, less than a quarter of that of the baseline loop, because candidates are benchmarked and diagnosed in \simulator{} and a GPU benchmark sweep covers a batch rather than a single candidate.
The additional candidates that asynchronous orchestration starts therefore rarely wait for the target server, and \texttt{\proposed{}\textsubscript{async}} accepts 41 candidates, compared with 17 for \texttt{\proposed{}\textsubscript{sync}}.
These results confirm that asynchronous orchestration shortens time-to-efficiency only when diagnosis does not occupy the target GPUs, which is what simulation-based diagnosis provides (\sect{subsect:aync_coder_orchestration}).

%% file: tex/related_work.tex
\section{Related Work}
\label{sect:related_work}\label{subsect:related}

{\bf Agent-driven LLM serving optimizations.}
The agent-driven optimization described in \sect{subsect:agent_optimization} builds on a rapidly growing body of work, which a recent survey catalogs at the kernel level~\cite{yu2026kernelsurvey}. At that level, KernelBench~\cite{kernelbench} established the task of rewriting PyTorch workloads as faster GPU kernels, and later benchmark suites added more robust verification~\cite{lange2025robustkbench} and kernels whose workloads are drawn from real LLM serving traces~\cite{flashinferbench2026}. The systems built on these suites differ mainly in how they organize the agent's search. Some refine kernels iteratively through evolutionary search~\cite{lange2025robustkbench,andrews2025gpukernelscientist,wiedemann2026kernelfoundry}, some divide planning, coding, and debugging among multiple agents~\cite{stark,astra,nagaitsev2026pike}, some feed hardware profiles back to the agent~\cite{zhang2025cudaforge}, some accumulate a memory of past optimizations for emerging accelerators~\cite{accelopt}, and some train the underlying model for kernel generation with reinforcement learning~\cite{baronio2026kevin}. All of these systems, however, accept or reject a kernel based on a microbenchmark run on the target accelerator, which does not reveal the kernel's effect on end-to-end serving. \proposed{} is orthogonal to the search algorithms that these systems use to optimize a kernel and can be applied on top of any of them, because it changes how a candidate kernel is benchmarked rather than how it is generated.

More recent work therefore asks agents to optimize, or even to build, the serving system itself. ISO-Bench~\cite{isobench2026} collects 54 optimization tasks from merged pull requests of vLLM and SGLang, while ServerBench~\cite{serverbench2026} has agents construct serving systems that are specialized for a given workload. These efforts benchmark candidates by running them on GPUs, and they are judged by the serving performance that the agents finally reach rather than by the time needed to reach it, i.e., their time-to-efficiency. \proposed{} is complementary to these systems, because it leaves the coding agents as they are and instead removes most GPU measurements from the loop by benchmarking and diagnosing candidates in a simulator.

{\bf LLM serving simulators.}\label{subsect:perf_models}
All of the systems above learn how a serving system performs by running it on GPUs, whereas a performance simulator, i.e., a model that predicts a system's performance without running it, obtains this information on CPUs. Simulators for LLM inference have been built to evaluate hardware designs~\cite{vserve2026,llmcompass,llmservingsim} and to compare serving configurations and scheduling policies at cluster scale~\cite{vidur}, and operators of large GPU fleets maintain calibrated simulators of their training and serving systems so that hardware and configurations can be planned without consuming accelerators~\cite{aiperf2026,simai2025,meta_inference_config2026,aiconfigurator_paper2026,dynosim2026,vtrain,vserve2026}. Because a prediction takes seconds to minutes on CPUs rather than hours on GPUs, such a simulator can sweep the plane of \fig{fig:pareto_frontier} and leave only a short list of frontier configurations to confirm on hardware~\cite{aiconfigurator_paper2026,dynosim2026}. These simulators, however, estimate the performance of a \emph{fixed} implementation and are used to choose among its configurations. Whether a simulator can take the place of GPU benchmarking and diagnosis in a loop where agents keep rewriting the serving stack's code, and how a provider should use it there, have not been examined. \proposed{} answers these questions by co-designing coding agents and a simulator in a closed loop (\sect{sect:proposed}), and because such simulators are not specific to \proposed{}, a provider can connect a simulator that it already maintains and trusts to this loop instead of building a new one from scratch.

%% file: tex/conclusion.tex
\section{Conclusion}
\label{sect:conclusion}

As the optimization of LLM serving stacks is increasingly delegated to AI
coding agents, writing code is no longer what limits time-to-efficiency. Our characterization
of agent-driven optimization of production serverless LLM endpoints shows that
time-to-efficiency is bottlenecked by the GPU time spent on benchmarking and
diagnosing the agents' candidates. Motivated by this key observation, we
propose \proposed{}, which replaces most GPU benchmarking and diagnosis with
simulation in \simulator{}, our LLM serving simulator. \proposed{} uses \simulator{} to explore the serving configuration space, to benchmark candidates, and to diagnose the serving system, which frees the GPUs from diagnosis and lets it orchestrate coders asynchronously. Compared with the baseline loop, which benchmarks every candidate on GPUs,
\proposed{} shortens time-to-efficiency by 1.1--15.1$\times$ (a geometric mean of 3.9$\times$) and, after the same 24-hour budget, achieves 1.06--1.11$\times$ higher per-GPU throughput.

%% file: refs.bib
@STRING{ISCA = {Proceedings of the International
               Symposium on Computer Architecture (ISCA)}}

@STRING{MICRO = {Proceedings of the International
                Symposium on Microarchitecture (MICRO)}}

@STRING{SOSP = {Proceedings of the ACM Symposium on
               Operating System Principles (SOSP)}}

@STRING{OSDI = {Proceedings of USENIX Symposium on
               Operating Systems Design and
               Implementation (OSDI)}}

@STRING{NeurIPS = {Proceedings of the International
                  Conference on Neural Information
                  Processing Systems (NeurIPS)}}

@STRING{ICLR = {Proceedings of the International
               Conference on Learning Representations
               (ICLR)}}

@STRING{ICML = {Proceedings of the International 
                Conference on Machine Learning
               (ICML)}}

@STRING{ATC_USENIX = {Proceedings of the USENIX Annual
              Technical Conference (ATC)}}

@STRING{ARXIV = {arxiv.org}}

@STRING{MLSYS = {Proceedings of Machine Learning and Systems (MLSys)}}

@STRING{IISWC = {Proceedings of the IEEE International
                Symposium on Workload Characterization
                (IISWC)}}

@STRING{ISPASS = {Proceedings of the IEEE International
                 Symposium on Performance Analysis of
                 Systems and Software (ISPASS)}}

@STRING{NSDI = {Proceedings of the USENIX Symposium on
               Networked Systems Design and
               Implementation (NSDI)}}

@misc{semianalysis_neocloud2024,
    title = {{AI} Neocloud Playbook and Anatomy},
    author = {Patel, Dylan and Nishball, Daniel},
    year = {2024},
  note = {\url{https://newsletter.semianalysis.com/p/ai-neocloud-playbook-and-anatomy}}
}

@misc{together_serverless,
    title = {Serverless Models: Overview},
    author = {{Together AI}},
    year = {2026},
  note = {\url{https://docs.together.ai/docs/serverless/overview}}
}

@misc{fireworks_serverless,
    title = {Serverless Inference Overview},
    author = {{Fireworks AI}},
    year = {2026},
  note = {\url{https://docs.fireworks.ai/serverless/overview}}
}

@misc{baseten_modelapis,
    title = {Model {APIs}},
    author = {{Baseten}},
    year = {2026},
  note = {\url{https://docs.baseten.co/inference/model-apis/overview}}
}

@misc{openrouter_quickstart,
    title = {Quickstart: Get Started with {OpenRouter}},
    author = {{OpenRouter}},
    year = {2026},
  note = {\url{https://openrouter.ai/docs/quickstart}}
}

@misc{vercel_aigateway,
    title = {Vercel {AI Gateway}: Models, Routing, and Observability},
    author = {{Vercel}},
    year = {2026},
  note = {\url{https://vercel.com/docs/ai-gateway}}
}

@misc{hf_inference_providers,
    title = {{Inference Providers}},
    author = {{Hugging Face}},
    year = {2026},
  note = {\url{https://huggingface.co/docs/inference-providers/index}}
}

@misc{fireworks_dayzero,
    title = {Distributed Inference Engine},
    author = {{Fireworks AI}},
    year = {2026},
  note = {\url{https://fireworks.ai/platform/disaggregated-inference-engine}}
}

@misc{semianalysis_valuecapture2026,
    title = {{AI} Value Capture: The Shift to Model Labs},
    author = {Nishball, Daniel and Patel, Dylan and Wen, Cheang Kang and Huang, Crystal and Kan, Max and Wang, Ray and Xie, Myron and Fong, Zane and Ee, Clara},
    year = {2026},
  note = {\url{https://newsletter.semianalysis.com/p/ai-value-capture-the-shift-to-model}}
}

@misc{openai_efficiency2026,
    title = {How {GPT}-5.6 Fuses Frontier Intelligence with Frontier Efficiency},
    author = {{OpenAI}},
    year = {2026},
  note = {\url{https://openai.com/index/gpt-5-6-frontier-intelligence-efficiency/}}
}

@misc{meta_rea2026,
    title = {{KernelEvolve}: How {Meta’s} {Ranking Engineer Agent} Optimizes {AI} Infrastructure},
    author = {Liao, Gang and Yetim, Yavuz and Xiao, Ruichao and Jiang, Zewei and Boinepalli, Raghav and Yadawad, Sheela and Li, Liyuan and Yan, Nathan and Mathews, Ajit and Tang, Chunqiang and Wu, Carole-Jean and Liu, Gaoxiang},
    year = {2026},
  note = {\url{https://engineering.fb.com/2026/04/02/developer-tools/kernelevolve-how-metas-ranking-engineer-agent-optimizes-ai-infrastructure/}}
}

@inproceedings{liao2026kernelevolve,
  title={{KernelEvolve: Scaling Agentic Kernel Coding for Heterogeneous AI Accelerators at Meta}},
  author = {Liao, Gang and Qin, Hongsen and Wang, Ying and Golden, Alicia and Kuchnik, Michael and Yetim, Yavuz and Xiao, Ruichao and Ang, Jia Jiunn and Fu, Chunli and He, Yihan and Hsia, Samuel and Jiang, Zewei and Levenstein, Roman and Li, Dianshi and Li, Liyuan and Mathews, Ajit and Puvvada, Varna and Shi, Feng and Yan, Nathan and Yu, Xiayu and Pashkevich, Uladzimir and Steiner, Matt and Wu, Carole-Jean and Liu, Gaoxiang},
  booktitle= ISCA,
  year={2026}
}

@misc{baseten_agentic_kernels2026,
    title = {Agentic Kernels in Production},
    author = {Li, Brian and Shahsavan, Faraz and Gupta, Pankaj},
    year = {2026},
  note = {\url{https://www.baseten.co/blog/agentic-kernels-in-production/}}
}

@misc{amd_hyperloom2026,
    title = {Hyperloom: Autonomous Agentic Inference Optimization for {AMD} {GPUs}},
    author = {Farooq, Arsalan and Grond, Marco and Roosta, Tanya and Basanta, Marilyn and Gu, Zhenyu and Barsoum, Emad},
    year = {2026},
  note = {\url{https://rocm.blogs.amd.com/software-tools-optimization/hyperloom/README.html}}
}

@misc{cursor_multiagent2026,
    title = {Speeding Up {GPU} Kernels by 38\% with a Multi-Agent System},
    author = {Lin, Wilson and Modi, Sahil and Zhang, Yuan and Lin, Edward},
    year = {2026},
  note = {\url{https://cursor.com/blog/multi-agent-kernels}}
}

@inproceedings{serverbench2026,
  title={{ServerBench}: How Can Your Coding Agents Design Specialized {LLM} Serving Systems?},
  author = {Zhong, Xinrui and Wang, Zhiming and Liu, Yongqi and Wang, Zhaohui and Yang, Shanglin and Li, Zikun and Chen, Zhuoming and Chen, Beidi},
  booktitle=ARXIV,
  year={2026}
}

@inproceedings{vidur,
  title={Vidur: A Large-Scale Simulation Framework for {LLM} Inference},
  author = {Agrawal, Amey and Kedia, Nitin and Mohan, Jayashree and Panwar, Ashish and Kwatra, Nipun and Gulavani, Bhargav S. and Ramjee, Ramachandran and Tumanov, Alexey},
  booktitle=MLSYS,
  year={2024}
}

@inproceedings{meta_inference_config2026,
  title={Optimizing Deployment Configurations for {LLM} Inference},
  author = {Cho, Sung Min and Lee, Jaewon and Tang, Chunqiang and Lee, Yejin and Jeong, Geonhwa and Agape, Anca and Batura, Scott and Boivin, Vincent and Chen, Stephen and Chen, Renfei and Chen, Sijia and Cui, Yan and Davis, Bradley and Deng, Zhaoxia and Egebo, Nick and El-Haraty, Emad and Estienne, Sebastien and Fang, Lu and Fromm, Josh and Ganapathy, Raj and Goswami, Vedanuj and Guo, Leo and Hu, Ye and Hua, Chenheli and Huang, Jianyu and Ibrahim, Aya and Jagannath, Niranjan and Jia, Hongyi and Kim, Changkyu and Li, Shikai and Liu, Brandon and Liu, Jiawen and Mathews, Ajit and Meng, Xiaozhu and Mihailescu, Vlad Tiberiu and Nagpal, Amit and Naumov, Maxim and Ostrowski, Michal and Ouyang, Jialin and Park, Jason and Pumma, Sarunya and Qi, Ye and Qi, Zixi and Reizenstein, Jeremy and Saha, Rajasi and Santhanam, Nandhini and Shu, Zhan and Silva, Ruan and Sizov, Grigory and Swenson, Jon Maltiel and Taylor, Brandon and Thi, Chris and Victoria, Adolfo and Wang, Yunfan and Wang, Pengchao and Wang, Wenchen and Wang, Xiaodong and Wasti, Bram and Xu, Wei and Yang, Qirui and Yang, Jingyi and Yuen, Hector and Zhang, Zhengyuan and Zhang, Jing and Zhen, Jenny and Zhou, Yanjun},
  booktitle=MLSYS,
  year={2026}
}

@inproceedings{aiconfigurator_paper2026,
  title={{AIConfigurator}: Lightning-Fast Configuration Optimization for Multi-Framework {LLM} Serving},
  author = {Xu, Tianhao and Liu, Yiming and Lu, Xianglong and Zhao, Yijia and Zhou, Xuting and Feng, Aichen and Chen, Yiyi and Shen, Yi and Zhou, Qin and Chen, Xumeng and Sherstyuk, Ilya and Li, Haorui and Thakkar, Rishi and Hamm, Ben and Li, Yuanzhe and Huang, Xue and Wu, Wenpeng and Shanbhag, Anish and Kim, Harry and Chen, Chuan and Lai, Junjie},
  booktitle=ARXIV,
  year={2026}
}

@misc{dynosim2026,
    title = {{DynoSim}: Simulating the {Pareto} Frontier},
    author = {Ding, Yongming and Pei, Rudy and Zhou, Hongkuan and Olson, Ryan and Flowers, Alec and Mailthody, Vikram Sharma},
    year = {2026},
  note = {\url{https://developer.nvidia.com/blog/dynosim-simulating-the-pareto-frontier}}
}

@misc{agentx,
    title = {{AgentX} benchmark datasets},
    author = {{SemiAnalysis}},
    year = {2026},

}

@misc{mlperf_llm_v5,
    title = {{MLPerf Inference} v5.0 Advances Language Model Capabilities for {GenAI}},
    author = {Atta-Fosu, Thomas and Chandra, Viraat and Hodak, Miro and Jiang, Zhihan and Nanjappa, Ashwin},
    year = {2025},
  note = {\url{https://mlcommons.org/2025/04/llm-inference-v5/}}
}

@misc{crusoe_inferencemax2025,
    title = {The New {AI} Benchmark: Unlocking Real-World Performance with {InferenceMAX} by {SemiAnalysis}},
    author = {Menard, Erwan},
    year = {2025},
  note = {\url{https://www.crusoe.ai/resources/blog/the-new-ai-benchmark-unlocking-real-world-performance-with-inferencemax-by-semianalysis}}
}

@misc{nvidia_codesign2026,
    title = {{AI} Model Co-Design: Hardware-Friendly {LLM} Design},
    author = {Borkar, Ritika and Bhatia, Nidhi and Gopireddy, Bhargava and Comly, Nick and Pharris, Brian and Demouth, Julien and Rouhani, Bita Darvish},
    year = {2026},
  note = {\url{https://developer.nvidia.com/blog/ai-model-co-design-hardware-friendly-llm-design/}}
}

@misc{nvidia_dynamo_moe2025,
    title = {How {NVIDIA} {GB200} {NVL72} and {NVIDIA} {Dynamo} Boost Inference Performance for {MoE} Models},
    author = {Mitra, Tiyasa and Borkar, Ritika and Elmeleegy, Amr and Kapasi, Ujval and Rouhani, Bita Darvish},
    year = {2025},
  note = {\url{https://developer.nvidia.com/blog/how-nvidia-gb200-nvl72-and-nvidia-dynamo-boost-inference-performance-for-moe-models/}}
}

@misc{inferencex,
    title = {{InferenceX}™, Open Source Inference Research Platform},
    author = {{SemiAnalysis}},
    year = {2026},
  note = {\url{https://github.com/SemiAnalysisAI/InferenceX}}
}

@misc{vllm_inferencemax2025,
    title = {{SemiAnalysis} {InferenceMAX}: {vLLM} and {NVIDIA} Accelerate {Blackwell} Inference},
    author = {{vLLM Team}},
    year = {2025},
  note = {\url{https://vllm.ai/blog/2025-10-09-blackwell-inferencemax}}
}

@inproceedings{vllm,
  title={Efficient Memory Management for Large Language Model Serving with {PagedAttention}},
  author = {Kwon, Woosuk and Li, Zhuohan and Zhuang, Siyuan and Sheng, Ying and Zheng, Lianmin and Yu, Cody Hao and Gonzalez, Joseph E. and Zhang, Hao and Stoica, Ion},
  booktitle=SOSP,
  year={2023}
}

@inproceedings{sglang,
	title = {{SGLang}: {Efficient} {Execution} of {Structured} {Language} {Model} {Programs}},
	booktitle = NeurIPS,
    author = {Zheng, Lianmin and Yin, Liangsheng and Xie, Zhiqiang and Sun, Chuyue and Huang, Jeff and Yu, Cody Hao and Cao, Shiyi and Kozyrakis, Christos and Stoica, Ion and Gonzalez, Joseph E. and Barrett, Clark and Sheng, Ying},
	year = {2024}
}

@inproceedings{orca,
	title = {Orca: {A} {Distributed} {Serving} {System} for {Transformer}-{Based} {Generative} {Models}},
	booktitle = OSDI,
    author = {Yu, Gyeong-In and Jeong, Joo Seong and Kim, Geon-Woo and Kim, Soojeong and Chun, Byung-Gon},
	year = {2022}
}

@inproceedings{DCPpaper,
    title = {Helix Parallelism: Rethinking Sharding Strategies for Interactive Multi-Million-Token {LLM} Decoding},
    booktitle = ARXIV,
    author = {Nidhi Bhatia and Ankit More and Ritika Borkar and Tiyasa Mitra and Ramon Matas and Ritchie Zhao and Maximilian Golub and Dheevatsa Mudigere and Brian Pharris and Bita Darvish Rouhani},
    year = {2025}
}

@inproceedings{distserve,
	title = {{DistServe}: {Disaggregating} {Prefill} and {Decoding} for {Goodput}-optimized {Large} {Language} {Model} {Serving}},
	booktitle = OSDI,
    author = {Zhong, Yinmin and Liu, Shengyu and Chen, Junda and Hu, Jianbo and Zhu, Yibo and Liu, Xuanzhe and Jin, Xin and Zhang, Hao},
	year = {2024}
}

@inproceedings{splitwise,
	title = {Splitwise: {Efficient} {Generative} {LLM} {Inference} {Using} {Phase} {Splitting}},
	booktitle = ISCA,
	author = {Patel, Pratyush and Choukse, Esha and Zhang, Chaojie and Shah, Aashaka and Goiri, Íñigo and Maleki, Saeed and Bianchini, Ricardo},
	year = {2024}
}

@inproceedings{eagle,
	title = {{EAGLE}-3: {Scaling} up {Inference} {Acceleration} of {Large} {Language} {Models} via {Training}-{Time} {Test}},
	booktitle = NeurIPS,
	author = {Li, Yuhui and Wei, Fangyun and Zhang, Chao and Zhang, Hongyang},
	year = {2025}
}

@inproceedings{deepseekv3,
    title = {{DeepSeek}-{V3} {Technical} {Report}},
    booktitle = ARXIV,
	author = {DeepSeek-AI and Liu, Aixin and Feng, Bei and Xue, Bing and Wang, Bingxuan and Wu, Bochao and Lu, Chengda and Zhao, Chenggang and Deng, Chengqi and Zhang, Chenyu and Ruan, Chong and Dai, Damai and Guo, Daya and Yang, Dejian and Chen, Deli and Ji, Dongjie and Li, Erhang and Lin, Fangyun and Dai, Fucong and Luo, Fuli and Hao, Guangbo and Chen, Guanting and Li, Guowei and Zhang, H. and Bao, Han and Xu, Hanwei and Wang, Haocheng and Zhang, Haowei and Ding, Honghui and Xin, Huajian and Gao, Huazuo and Li, Hui and Qu, Hui and Cai, J. L. and Liang, Jian and Guo, Jianzhong and Ni, Jiaqi and Li, Jiashi and Wang, Jiawei and Chen, Jin and Chen, Jingchang and Yuan, Jingyang and Qiu, Junjie and Li, Junlong and Song, Junxiao and Dong, Kai and Hu, Kai and Gao, Kaige and Guan, Kang and Huang, Kexin and Yu, Kuai and Wang, Lean and Zhang, Lecong and Xu, Lei and Xia, Leyi and Zhao, Liang and Wang, Litong and Zhang, Liyue and Li, Meng and Wang, Miaojun and Zhang, Mingchuan and Zhang, Minghua and Tang, Minghui and Li, Mingming and Tian, Ning and Huang, Panpan and Wang, Peiyi and Zhang, Peng and Wang, Qiancheng and Zhu, Qihao and Chen, Qinyu and Du, Qiushi and Chen, R. J. and Jin, R. L. and Ge, Ruiqi and Zhang, Ruisong and Pan, Ruizhe and Wang, Runji and Xu, Runxin and Zhang, Ruoyu and Chen, Ruyi and Li, S. S. and Lu, Shanghao and Zhou, Shangyan and Chen, Shanhuang and Wu, Shaoqing and Ye, Shengfeng and Ye, Shengfeng and Ma, Shirong and Wang, Shiyu and Zhou, Shuang and Yu, Shuiping and Zhou, Shunfeng and Pan, Shuting and Wang, T. and Yun, Tao and Pei, Tian and Sun, Tianyu and Xiao, W. L. and Zeng, Wangding and Zhao, Wanjia and An, Wei and Liu, Wen and Liang, Wenfeng and Gao, Wenjun and Yu, Wenqin and Zhang, Wentao and Li, X. Q. and Jin, Xiangyue and Wang, Xianzu and Bi, Xiao and Liu, Xiaodong and Wang, Xiaohan and Shen, Xiaojin and Chen, Xiaokang and Zhang, Xiaokang and Chen, Xiaosha and Nie, Xiaotao and Sun, Xiaowen and Wang, Xiaoxiang and Cheng, Xin and Liu, Xin and Xie, Xin and Liu, Xingchao and Yu, Xingkai and Song, Xinnan and Shan, Xinxia and Zhou, Xinyi and Yang, Xinyu and Li, Xinyuan and Su, Xuecheng and Lin, Xuheng and Li, Y. K. and Wang, Y. Q. and Wei, Y. X. and Zhu, Y. X. and Zhang, Yang and Xu, Yanhong and Xu, Yanhong and Huang, Yanping and Li, Yao and Zhao, Yao and Sun, Yaofeng and Li, Yaohui and Wang, Yaohui and Yu, Yi and Zheng, Yi and Zhang, Yichao and Shi, Yifan and Xiong, Yiliang and He, Ying and Tang, Ying and Piao, Yishi and Wang, Yisong and Tan, Yixuan and Ma, Yiyang and Liu, Yiyuan and Guo, Yongqiang and Wu, Yu and Ou, Yuan and Zhu, Yuchen and Wang, Yuduan and Gong, Yue and Zou, Yuheng and He, Yujia and Zha, Yukun and Xiong, Yunfan and Ma, Yunxian and Yan, Yuting and Luo, Yuxiang and You, Yuxiang and Liu, Yuxuan and Zhou, Yuyang and Wu, Z. F. and Ren, Z. Z. and Ren, Zehui and Sha, Zhangli and Fu, Zhe and Xu, Zhean and Huang, Zhen and Zhang, Zhen and Xie, Zhenda and Zhang, Zhengyan and Hao, Zhewen and Gou, Zhibin and Ma, Zhicheng and Yan, Zhigang and Shao, Zhihong and Xu, Zhipeng and Wu, Zhiyu and Zhang, Zhongyu and Li, Zhuoshu and Gu, Zihui and Zhu, Zijia and Liu, Zijun and Li, Zilin and Xie, Ziwei and Song, Ziyang and Gao, Ziyi and Pan, Zizheng},
	year = {2025}
}

@inproceedings{lmcache,
	title = {{LMCache}: {An} {Efficient} {KV} {Cache} {Layer} for {Enterprise}-{Scale} {LLM} {Inference}},
	booktitle = ARXIV,
	author = {Liu, Yuhan and Cheng, Yihua and Yao, Jiayi and An, Yuwei and Chen, Xiaokun and Feng, Shaoting and Huang, Yuyang and Shen, Samuel and Zhang, Rui and Du, Kuntai and Jiang, Junchen},
	year = {2025}
}

@inproceedings{sweagent,
	title = {{SWE}-agent: {Agent}-{Computer} {Interfaces} {Enable} {Automated} {Software} {Engineering}},
	booktitle = NeurIPS,
	author = {Yang, John and Jimenez, Carlos and Wettig, Alexander and Lieret, Kilian and Yao, Shunyu and Narasimhan, Karthik and Press, Ofir},
	year = {2024}
}

@inproceedings{react,
	title = {{ReAct}: {Synergizing} {Reasoning} and {Acting} in {Language} {Models}},
    booktitle = ICLR,
	author = {Shunyu Yao and Jeffrey Zhao and Dian Yu and Nan Du and Izhak Shafran and Karthik Narasimhan and Yuan Cao},
	year = {2023}
}

@inproceedings{kernelbench,
  title={{KernelBench}: Can {LLMs} Write Efficient {GPU} Kernels?},
  author = {Ouyang, Anne and Guo, Simon and Arora, Simran and Zhang, Alex L. and Hu, William and Ré, Christopher and Mirhoseini, Azalia},
  booktitle=ICML,
  year={2025}
}

@inproceedings{stark,
  title={{STARK}: Strategic Team of Agents for Refining Kernels},
  author = {Dong, Juncheng and Yang, Yang and Liu, Tao and Wang, Yang and Qi, Feng and Tarokh, Vahid and Rangadurai, Kaushik and Yang, Shuang},
  booktitle=ARXIV,
  year={2025}
}

@inproceedings{astra,
  title={Astra: A Multi-Agent System for {GPU} Kernel Performance Optimization},
  author = {Wei, Anjiang and Sun, Tianran and Seenichamy, Yogesh and Song, Hang and Ouyang, Anne and Mirhoseini, Azalia and Wang, Ke and Aiken, Alex},
  booktitle=ARXIV,
  year={2025}
}

@inproceedings{accelopt,
  title={{AccelOpt}: A Self-Improving {LLM} Agentic System for {AI} Accelerator Kernel Optimization},
  author = {Zhang, Genghan and Zhu, Shaowei and Wei, Anjiang and Song, Zhenyu and Nie, Allen and Jia, Zhen and Vijaykumar, Nandita and Wang, Yida and Olukotun, Kunle},
  booktitle=MLSYS,
  year={2026}
}

@inproceedings{lina,
	title = {Accelerating {Distributed} {MoE} {Training} and {Inference} with {Lina}},
    booktitle = ATC_USENIX,
	author = {Li, Jiamin and Jiang, Yimin and Zhu, Yibo and Wang, Cong and Xu, Hong},
	year = {2023}
}

@inproceedings{fastkernels,
    title = {{FastKernels}: Benchmarking {GPU} Kernel Generation in Production},
    booktitle = ARXIV,
    author = {Oliaro, Gabriele and Lee, Jaeseong and Fu, Yichao and Su, Zhaoyuan and Lu, Owen and Vennam, Sreeram and Jiang, May and Wang, Junli and Zhang, Hao and Jia, Zhihao and Rajbhandari, Samyam},
    year = {2026}
}

@inproceedings{kernelopt,
    title = {{KernelOPT}: Dispatch-Aware Agentic Search for {GPU} Kernel Optimization},
    booktitle = ARXIV,
    author = {Poddar, Aheli and Prasad, Sanskar and Samanta, Arindam and Chakraborty, Subha and Goyal, Vishal and Rathaur, Rohit Singh},
    year = {2026}
}

@inproceedings{yu2026kernelsurvey,
  title={Towards Automated Kernel Generation in the Era of {LLMs}},
  author = {Yu, Yang and Zang, Peiyu and Tsai, Chi Hsu and Wu, Haiming and Shen, Yixin and Zhang, Jialing and Wang, Haoyu and Xiao, Zhiyou and Shi, Jingze and Luo, Yuyu and Zhang, Wentao and Men, Chunlei and Liu, Guang and Lin, Yonghua},
  booktitle={Proceedings of the International Joint
  Conference on Artificial Intelligence (IJCAI)},
  year={2026}
}

@inproceedings{lange2025robustkbench,
  title={Towards Robust Agentic {CUDA} Kernel Benchmarking, Verification, and Optimization},
  author = {Lange, Robert Tjarko and Sun, Qi and Prasad, Aaditya and Faldor, Maxence and Tang, Yujin and Ha, David},
  booktitle=ARXIV,
  year={2025}
}

@inproceedings{flashinferbench2026,
  title={{FlashInfer-Bench}: Building the Virtuous Cycle for {AI}-driven {LLM} Systems},
  author = {Xing, Shanli and Zhai, Yiyan and Jiang, Alexander and Dong, Yixin and Wu, Yong and Ye, Zihao and Ruan, Charlie F. and Huang, Yingyi and Zhang, Yineng and Yin, Liangsheng and Bayyapu, Aksara and Ceze, Luis and Chen, Tianqi},
  booktitle=MLSYS,
  year={2026}
}

@inproceedings{andrews2025gpukernelscientist,
  title={{GPU} {Kernel Scientist}: An {LLM}-Driven Framework for Iterative Kernel Optimization},
  author = {Andrews, Martin and Witteveen, Sam},
  booktitle=ARXIV,
  year={2025}
}

@inproceedings{wiedemann2026kernelfoundry,
  title={{KernelFoundry}: Hardware-aware evolutionary {GPU} kernel optimization},
  author={Wiedemann, Nina and Leboutet, Quentin and Paulitsch, Michael and Wofk, Diana and Ummenhofer, Benjamin},
  booktitle=ICML,
  year={2026}
}

@inproceedings{nagaitsev2026pike,
  title={Optimizing {PyTorch} Inference with {LLM}-Based Multi-Agent Systems},
  author = {Nagaitsev, Kirill and Grbcic, Luka and Williams, Samuel and Iancu, Costin},
  booktitle=MLSYS,
  year={2026}
}

@inproceedings{zhang2025cudaforge,
  title={{CudaForge}: An Agent Framework with Hardware Feedback for {CUDA} Kernel Optimization},
  author={Zhang, Zijian and Wang, Rong and Li, Shiyang and Luo, Yuebo and Hong, Mingyi and Ding, Caiwen},
  booktitle=ARXIV,
  year={2025}
}

@inproceedings{baronio2026kevin,
  title={Kevin: Multi-Turn {RL} for Generating {CUDA} Kernels},
  author = {Baronio, Carlo and Marsella, Pietro and Pan, Ben and Guo, Simon and Alberti, Silas},
  booktitle=ICLR,
  year={2026}
}

@inproceedings{isobench2026,
  title={{ISO-Bench}: Can Coding Agents Optimize Real-World Inference Workloads?},
  author = {Nangia, Ayush and Mishra, Shikhar and Gokrani, Aman and Chopra, Paras},
  booktitle=ARXIV,
  year={2026}
}

@inproceedings{llmcompass,
  title={{LLMCompass}: Enabling Efficient Hardware Design for Large Language Model Inference},
  author = {Zhang, Hengrui and Ning, August and Prabhakar, Rohan Baskar and Wentzlaff, David},
  booktitle=ISCA,
  year={2024}
}

@inproceedings{llmservingsim,
  title={{LLMServingSim}: A {HW}/{SW} Co-Simulation Infrastructure for {LLM} Inference Serving at Scale},
  author = {Cho, Jaehong and Kim, Minsu and Choi, Hyunmin and Heo, Guseul and Park, Jongse},
  booktitle=IISWC,
  year={2024}
}

@inproceedings{llmservingsim2,
  title={{LLMServingSim} 2.0: A Unified Simulator for Heterogeneous and Disaggregated {LLM} Serving Infrastructure},
  author = {Cho, Jaehong and Choi, Hyunmin and Heo, Guseul and Park, Jongse},
  booktitle=ISPASS,
  year={2026}
}

@inproceedings{aiperf2026,
  title={{AIPerf}: {Meta’s} Performance Modeling and {HW/SW} Co-design Framework for Datacenter-Scale {LLM} Training},
  author = {Park, Pyeongsu and Lim, Hwijoon and Kim, Seah and Ozdal, Mustafa and Lin, Yi-Chien and Siu, Kevin and Yalamanchili, Pavan and Ghaemmaghami, Benjamin and Gu, Yufeng and Park, Jongsoo and Coburn, Joel and Baumstarck, Paul and Kanaujia, Shobhit and Tang, Chunqiang},
  booktitle=MICRO,
  year={2026}
}

@inproceedings{simai2025,
  title={{SimAI}: Unifying Architecture Design and Performance Tuning for Large-Scale Large Language Model Training with Scalability and Precision},
  author = {Wang, Xizheng and Li, Qingxu and Xu, Yichi and Lu, Gang and Li, Dan and Chen, Li and Zhou, Heyang and Zheng, Linkang and Zhang, Sen and Zhu, Yikai and Liu, Yang and Zhang, Pengcheng and Qian, Kun and He, Kunling and Gao, Jiaqi and Zhai, Ennan and Cai, Dennis and Fu, Binzhang},
  booktitle=NSDI,
  year={2025}
}

@inproceedings{open_llm_survey,
  title = {A Comprehensive Overview of Large Language Models},
  booktitle = ARXIV,
  author = {Naveed, Humza and Khan, Asad Ullah and Qiu, Shi and Saqib, Muhammad and Anwar, Saeed and Usman, Muhammad and Akhtar, Naveed and Barnes, Nick and Mian, Ajmal},
  year = {2024}
}

@misc{vllm_day0,
    title = {Day 0 Support for {Qwen3.8}-{2.4T}-{A95B} on {vLLM}},
    author = {vLLM Team and Inferact},
    year = {2026},
  note = {\url{https://vllm.ai/blog/2026-08-12-qwen3.8}}
}

@misc{sglang_day0,
    title = {Day-0 Model Support},
    author = {SGLang},
    year = {2026},
  note = {\url{https://www.sglang.io/models}}
}

@misc{vllm_example_1,
    title = {Hyper Optimizing {Kimi} {K3} in {vLLM}: Moving Fast and Responsibly},
    author = {Chislett, Benjamin},
    year = {2026},
  note = {\url{https://vllm.ai/blog/2026-10-16-kimi-k3-megakernel}}
}

@misc{vllm_example_2,
    title = {{Kimi} {K3} Performance Optimizations in {vLLM}: The Road to 2.8× Throughput},
    author = {Ye, Wentao and Guo, Canlin and Zhu, Yongye and Zhu, Jiangyun and Huang, Ziming and Zhao, Wei and Goin, Michael and Li, Jie},
    year = {2026},

}

@misc{agentskills,
    title = {Equipping agents for the real world with {Agent Skills}},
    author = {Zhang, Barry and Lazuka, Keith and Murag, Mahesh},
    year = {2025},
  note = {\url{https://www.anthropic.com/research/equipping-agents-for-the-real-world-with-agent-skills}}
}

@misc{vllm_recipe,
    title = {{vLLM} {Recipes} — Deploy any model on any hardware with {vLLM}},
    author = {vLLM},
    year = {2026},
  note = {\url{https://recipes.vllm.ai/}}
}

@misc{torch_profiler,
    title = {{PyTorch} {Profiler}},
    author = {Raikundalia, Shivam},
    year = {2026},
  note = {\url{https://docs.pytorch.org/tutorials/recipes/recipes/profiler_recipe.html#pytorch-profiler}}
}

@misc{nvidia_nsys,
    title = {{NVIDIA} {Nsight Systems}},
    author = {NVIDIA},
    year = {2026},

}

@misc{claude_code,
    title = {Overview - {Claude Code Docs}},
    author = {Anthropic},
    year = {2026},
  note = {\url{https://code.claude.com/docs/en/overview}}
}

@misc{opus_5.5,
    title = {Introducing {Claude Opus} 5.5},
    author = {Anthropic},
    year = {2026},
  note = {\url{https://www.anthropic.com/claude-opus-5-5}}
}

@inproceedings{cobbe2021gsm8k,
  title = {Training Verifiers to Solve Math Word Problems},
  booktitle = ARXIV,
  author = {Cobbe, Karl and Kosaraju, Vineet and Bavarian, Mohammad and Chen, Mark and Jun, Heewoo and Kaiser, Lukasz and Plappert, Matthias and Tworek, Jerry and Hilton, Jacob and Nakano, Reiichiro and Hesse, Christopher and Schulman, John},
  year = {2021}
}

@inproceedings{hendrycks2021mmlu,
	title = {Measuring {Massive} {Multitask} {Language} {Understanding}},
    booktitle = ICLR,
	author = {Hendrycks, Dan and Burns, Collin and Basart, Steven and Zou, Andy and Mazeika, Mantas and Song, Dawn and Steinhardt, Jacob},
	year = {2021},
}

@inproceedings{glm_5.3_flash,
    title={{GLM}-5: from Vibe Coding to Agentic Engineering},
    booktitle=ARXIV,
    author={GLM-5-Team and Aohan Zeng and Xin Lv and Zhenyu Hou and Zhengxiao Du and Qinkai Zheng and Bin Chen and Da Yin and Chendi Ge and Chenghua Huang and Chengxing Xie and Chenzheng Zhu and Congfeng Yin and Cunxiang Wang and Gengzheng Pan and Hao Zeng and Haoke Zhang and Haoran Wang and Huilong Chen and Jiajie Zhang and Jian Jiao and Jiaqi Guo and Jingsen Wang and Jingzhao Du and Jinzhu Wu and Kedong Wang and Lei Li and Lin Fan and Lucen Zhong and Mingdao Liu and Mingming Zhao and Pengfan Du and Qian Dong and Rui Lu and Shuang-Li and Shulin Cao and Song Liu and Ting Jiang and Xiaodong Chen and Xiaohan Zhang and Xuancheng Huang and Xuezhen Dong and Yabo Xu and Yao Wei and Yifan An and Yilin Niu and Yitong Zhu and Yuanhao Wen and Yukuo Cen and Yushi Bai and Zhongpei Qiao and Zihan Wang and Zikang Wang and Zilin Zhu and Ziqiang Liu and Zixuan Li and Bojie Wang and Bosi Wen and Can Huang and Changpeng Cai and Chao Yu and Chen Li and Chengwei Hu and Chenhui Zhang and Dan Zhang and Daoyan Lin and Dayong Yang and Di Wang and Ding Ai and Erle Zhu and Fangzhou Yi and Feiyu Chen and Guohong Wen and Hailong Sun and Haisha Zhao and Haiyi Hu and Hanchen Zhang and Hanrui Liu and Hanyu Zhang and Hao Peng and Hao Tai and Haobo Zhang and He Liu and Hongwei Wang and Hongxi Yan and Hongyu Ge and Huan Liu and Huanpeng Chu and Jia'ni Zhao and Jiachen Wang and Jiajing Zhao and Jiamin Ren and Jiapeng Wang and Jiaxin Zhang and Jiayi Gui and Jiayue Zhao and Jijie Li and Jing An and Jing Li and Jingwei Yuan and Jinhua Du and Jinxin Liu and Junkai Zhi and Junwen Duan and Kaiyue Zhou and Kangjian Wei and Ke Wang and Keyun Luo and Laiqiang Zhang and Leigang Sha and Liang Xu and Lindong Wu and Lintao Ding and Lu Chen and Minghao Li and Nianyi Lin and Pan Ta and Qiang Zou and Rongjun Song and Ruiqi Yang and Shangqing Tu and Shangtong Yang and Shaoxiang Wu and Shengyan Zhang and Shijie Li and Shuang Li and Shuyi Fan and Wei Qin and Wei Tian and Weining Zhang and Wenbo Yu and Wenjie Liang and Xiang Kuang and Xiangmeng Cheng and Xiangyang Li and Xiaoquan Yan and Xiaowei Hu and Xiaoying Ling and Xing Fan and Xingye Xia and Xinyuan Zhang and Xinze Zhang and Xirui Pan and Xu Zou and Xunkai Zhang and Yadi Liu and Yandong Wu and Yanfu Li and Yidong Wang and Yifan Zhu and Yijun Tan and Yilin Zhou and Yiming Pan and Ying Zhang and Yinpei Su and Yipeng Geng and Yong Yan and Yonglin Tan and Yuean Bi and Yuhan Shen and Yuhao Yang and Yujiang Li and Yunan Liu and Yunqing Wang and Yuntao Li and Yurong Wu and Yutao Zhang and Yuxi Duan and Yuxuan Zhang and Zezhen Liu and Zhengtao Jiang and Zhenhe Yan and Zheyu Zhang and Zhixiang Wei and Zhuo Chen and Zhuoer Feng and Zijun Yao and Ziwei Chai and Ziyuan Wang and Zuzhou Zhang and Bin Xu and Minlie Huang and Hongning Wang and Juanzi Li and Yuxiao Dong and Jie Tang},
    year={2026},
}

@inproceedings{deepseekv41flash,
	title = {{DeepSeek-V4.1-Flash}: Pushing the Limits of {KV} Cache Compression},
    booktitle = ARXIV,
    author = {DeepSeek-AI and Xu, Anyi and Li, B. and Lin, Bangcai and Xue, Bing and Xian, BingCheng and Xu, Bingzheng and Wu, Bochao and Zhang, Bowei and Deng, Boyi and Yu, C. C. and Jin, Chao and Lin, Chaofan and Dong, Chen and Wang, Chenbing and Feng, Chenfan and Lu, Chengda and Zhao, Chenggang and Deng, Chengqi and Zhang, Chengyuan and Xu, Chenhao and Zhao, Chenqi and Shao, Chenze and Wang, Chuhao and Zhang, Chuqi and Dai, Damai and Yang, Dejian and Chen, Deli and Huang, Di and Wu, Di and Li, Donghao and Li, Erhang and Fu, Eric and Zhou, F. and Zhou, Fangwei and Lin, Fangyun and Yuan, Fangzhou and Xia, Feiyu and Dai, Fucong and Hao, Guangbo and Li, Guanglin and Chen, Guanting and Cao, Guoai and Fan, Guofan and Meng, Guolai and Li, Guowei and Zhang, Haichuan and Ma, Haiyang and Shen, Haiyang and Li, Han and Yu, Han and Zhang, Han and Deng, Hangyuan and Xu, Hanwei and Xu, Hanxiang and Zhong, Hanxun and Guo, Hao and Jiang, Hao and Li, Hao and Qin, Hao and Wen, Haodong and Liang, Haofen and Huang, Haofeng and Liu, Haohua and Zhang, Haoling and Luo, Haoming and Yang, Haoran and Xu, Haotian and Yuan, Haotian and Huang, Haoting and Luo, Haowen and Cai, Haoyang and Chen, Haoyu and Ji, Haozhe and Zhang, Hengran and Wang, Hengrui and Wu, Hengxu and Ding, Honghui and Tang, Hongxuan and Wang, Huadong and Cao, Huanqi and Gao, Huazuo and Qu, Hui and Zeng, Hui and Yang, J. and Jin, J. H. and Zhang, J. H. and Zou, J. X. and Yu, Jia and Zhou, Jiahui and Chen, Jiajun and Huang, Jialiang and Zhao, Jialin and Tang, Jiamin and Zhou, Jian and Tong, Jianan and Li, Jianwen and Zhu, Jiaqi and Wang, Jiarui and Ye, Jiasheng and Li, Jiashi and Xu, Jiaxin and Ding, Jiaying and Lu, Jibai and Hu, Jiewen and Yan, Jin and Zhai, Jincheng and Chen, Jingchang and Hu, Jingcheng and Zhou, Jingli and Xu, Jingsheng and Xiang, Jingting and Yun, Jingyan and Yuan, Jingyang and Cheng, Jingyuan and Zhu, Jinhua and Wang, Jinpeng and Chen, Jinyi and Hu, Jinyi and Yu, Jiping and Guo, Jueliang and Pei, Junbo and Sun, Junbo and Jiang, Junguang and Qiu, Junjie and Zhou, Junkang and Liu, Junqi and Li, Junren and Li, Junxian and Song, Junxiao and Guo, Junyi and Dong, Kai and Chen, Kaifeng and Gao, Kaige and Guan, Kang and Yuan, Kangdong and Hong, Ke and Xu, Ke and Zhao, Kefan and Ji, Kexin and Zhang, Kexin and Zhou, Kexing and Yu, Kuai and Zhang, Lan and Wang, Lean and Zhang, Lecong and Wang, Lei and Gao, Letian and Zhao, Liang and Xu, Liansheng and Guo, Lihua and Luo, Lingxiao and Fu, Lingyue and Deng, Litao and Wang, Litong and Zhang, Liyue and Chen, Longhao and Chen, Lu and Huang, Luotian and Ma, Luyao and Wang, Luyao and Di, M. S. and Mei, Max and Ye, Menghao and Cui, Miao and Zhang, Mingchuan and Zhang, Minghua and Tang, Minghui and Zhang, Mingjing and Wei, Mingqi and Chen, Mingshu and Liu, Mingxing and Zhou, Mingxu and Xu, Mingyu and Yang, Mingyu and Wang, Mingze and Chen, Muyang and Shentu, Ni and Wang, Ning and Ning, Niufang and Huang, Panpan and Cong, Peixin and Wang, Peiyi and Xin, Peiyuan and Ren, Pengfei and Yan, Pengfei and Zhang, Pengle and Kang, Qi and Tang, Qi and Wang, Qiancheng and Li, Qiang and Zhu, Qihao and Li, Qingyang and Chen, Qinyu and Du, Qiushi and Guo, Qizhou and Xu, Rongxian and Ding, Rui and Hu, Rui and Tian, Rui and Yu, Rui and Zhu, Ruidong and Xu, Ruifan and Yang, Ruihan and Xia, Ruihang and Lu, Ruijie and Geng, Ruilin and Hong, Ruipeng and Ge, Ruiqi and Zhang, Ruisong and Sun, Ruize and Pan, Ruizhe and Wang, Runji and Chen, Runqian and Xu, Runxin and Tian, Ruohong and Shen, Ruomeng and Zhang, Ruoyu and X, Ryan and Liu, S. H. and Lu, Shanghao and Zhou, Shangyan and Chen, Shanhuang and Cai, Shaofei and Nie, Shaoheng and Chen, Shaoyuan and Hu, Shengding and Lin, Shengkai and Ran, Shengwen and Liu, Shengyu and Jia, Shengyuan and Bai, Shi and Feng, Shi and Xu, Shicheng and Liu, Shichun and Hu, Shiqiang and Ma, Shirong and Wang, Shiyu and Feng, Shiyuan and Gong, Shufan and Lin, Shuhan and Yu, Shuiping and Zhou, Shunfeng and Yang, Shuo and Wang, Shuomeng and Guo, Shuting and Pan, Shuting and Yu, Shuying and Cao, Sinuo and Lin, Siyi and Chen, Sizhe and Chen, Songyang and Zhou, Songyang and Ni, Tao and Yun, Tao and Jin, Tian and Pei, Tian and Ye, Tian and Lin, Tianle and Ji, Tianran and Cui, Tianyi and Yue, Tianyuan and Yu, Tingting and Xiong, Tongrui and Zeng, Wangding and Liu, Wei and Zhang, Wei and Xu, Weibin and Zeng, Weihao and Zhao, Weilin and Liu, Wen and Liang, Wenfeng and Pang, Wenjie and Luo, Wenjing and Yao, Wenjing and Gao, Wenjun and Shao, Wenkai and Yang, Wenkai and Zhang, Wenli and Wang, Wenlu and Huang, Wenlve and Yan, Wenqian and Zhang, Wentao and Gao, Xi and He, Xiang and Li, Xiang and Li, Xiangli and Wang, Xiangwen and Zhang, Xiangying and Wei, Xiankui and Bi, Xiao and Liu, Xiaodong and Wang, Xiaohan and Qu, Xiaojian and Chen, Xiaokang and Zhang, Xiaokang and Nie, Xiaotao and Zou, Xiaoyao and Li, Xiaoyuan and Guo, Xicheng and Chu, Xieting and Cheng, Xin and Liu, Xin and Xie, Xin and Xu, Xinbo and Liu, Xingchao and Liu, Xingchen and Yu, Xingkai and Li, Xingyou and Yao, Xintong and Chen, Xinyang and Jiang, Xinyong and Yang, Xinyu and Yang, Xinyu and Chen, Xu and Wang, Xuanyu and Zhong, Xubei and Su, Xuecheng and Liu, Xuejie and Lin, Xuheng and Fan, Xujie and Zhao, Xuncheng and Fu, Xuwei and Yan, Y. C. and Jiang, Y. H. and Wu, Y. T. and M, Y. W. and Wang, Y. Z. and Gao, Yafei and Yang, Yang and Zhang, Yang and Ma, Yanru and Huang, Yanwen and Li, Yao and Li, Yao and Meng, Yao and Zhao, Yao and Sun, Yaofeng and Wang, Yaohui and Ye, Yaoyang and Yin, Yehang and Wu, Yexinrui and Qian, Yi and Tao, Yi and Yu, Yi and Zhang, Yichao and Jiang, Yichen and Wang, Yicheng and Ding, Yifan and Shi, Yifan and Peng, Yifeng and Zhai, Yifeng and Wu, Yijia and Xiong, Yiliang and Wang, Yilun and He, Ying and Zhou, Ying and Luo, Yingjia and Zhong, Yinmin and Wang, Yiping and Wang, Yisong and Zhang, Yixiang and Chen, Yixiao and Tan, Yixuan and Wei, Yixuan and Ma, Yiyang and Yang, Yiyao and Liu, Yiyuan and Cai, Yizai and Wei, Yizhen and Wang, Yizhi and Yang, Yonglun and Zhuo, Yongqi and Guo, Yongqiang and Wu, Yongtong and Wu, Yu and Zhang, Yu and Bian, Yuan and Cheng, Yuan and Ou, Yuan and Sun, Yuan and Xu, Yuanfan and Sun, Yuanhang and Li, Yuanhao and Liu, Yuchen and Yao, Yuchen and Han, Yudong and Wang, Yuduan and Wu, Yuhan and Meng, Yuhao and Zou, Yuheng and Li, YuKun and Wang, Yunchuan and Xiao, Yunfan and Xiong, Yunfan and Chen, Yupeng and Cao, Yuqian and Wang, Yuqian and Chen, Yuqing and Zhang, Yushun and Lin, Yutong and Xiao, Yuwei and Gu, Yuxian and Chen, Yuxiang and Huang, Yuxiang and Luo, Yuxiang and You, Yuxiang and Chen, Yuxin and Xiang, Yuxin and Liu, Yuxuan and Zhou, Yuxuan and Zhou, Yuyang and Guo, Yuzhe and Huang, Yuzhen and Bai, Yuzhuo and Z, Z. Y. and Ni, Zanlin and Wang, Zehao and Zhao, Zehua and Ren, Zehui and Zhao, Zejun and Sha, Zhangli and Wang, Zhanying and Zhang, Zhaochen and Du, Zhaoshuai and Fu, Zhe and Xu, Zhean and Xie, Zhenda and Liu, Zheng and Zhang, Zhengyan and Dong, Zhenhua and Hao, Zhewen and Wang, Zhibang and Gou, Zhibin and Ma, Zhicheng and Li, Zhihao and Shao, Zhihong and Huang, Zhihuan and Li, Zhijie and Lu, Zhirui and Huang, Zhixian and Chen, Zhixuan and Chen, Zhixuan and Pan, Zhixuan and Wu, Zhiyu and Ren, Zhizhou and He, Zhu and Li, Zhuoshu and Zhang, Zhuping and Xu, Zian and Wang, Zihao and Gu, Zihui and Zhu, Zijia and Zhang, Zili and Li, Zilin and Hou, Zilong and Lyu, Zilong and Wang, Ziqiao and Xie, Ziwei and Zhang, Ziya and Gao, Ziyi and Pan, Zizheng and Li, Zonglin and Yao, Zongqing and Chen, Zui and Wu, Zuofan and Ling, Chenchen and Hou, Chengyu and Chen, Chong and Li, D. and Qi, Di and Ji, Dongjie and Wei, Fang and Xia, Fanyi and Xie, Fei and Tan, Feiyi and Guo, Hailong and Zhai, Haiyan and Zhou, Hui and Tan, Huihui and Li, Huijie and Luo, Jia and Song, Jia and Cai, Jialu and Liang, Jian and Zhou, Jiangting and Gao, Jiaqi and Shao, Jiayi and Chen, Jie and Yang, Jieyu and Chen, Jin and Zhang, Jingde and Zhou, Jingzi and Wang, Jinqian and Liu, Jinyang and Sun, JinZhao and Ling, Junhua and Zheng, Junmin and Yang, Kaicheng and Xu, Ke and Su, Le and Xia, Leyi and Ding, Liangfeng and Zhuo, Lin and Ma, Linwang and Zhu, Linyan and Cai, Liyu and Yao, Luqi and Zhang, M. K. and Li, Meng and Lin, Miao and Wang, Miaojun and Zhang, Min and Li, Mingming and Wang, Mingming and Yin, Mingze and Han, Minmin and Cao, Nan and Wang, Ning and Ma, Ningxin and Wang, Panpan and Lin, Peihan and Sun, Peng and Zhang, Peng and Ying, Qian and Xiang, Qiang and Wang, Qiao and Mao, Qingmiao and Jiang, Qiwei and Jin, Rongli and Chen, Ruyi and Tao, Sha and Sun, Shangmian and Wu, Shaoqing and Zou, Shichao and Lei, Si and Zhang, Tianyang and Sun, Tianyu and Yin, Tingting and Xiao, W. L. and An, Wei and Li, Wei and Wang, Wei and Lin, Weiwei and Hou, Wenqing and Lin, X. and Meng, Xiangfei and Huang, Xianzhu and Peng, Xiao and Li, Xiaoqian and Zhang, Xiaoting and Sun, Xiaowen and Wang, Xiaoxiang and Ye, Xiaoyu and Zhang, Xinrou and Zhang, Xinyu and Cao, Xue and Chen, Xueyin and Zhou, Yanan and Xu, Yanhong and Xia, Yao and Xu, Yao and Shao, Yi and Zhang, Yihong and Ma, Yiling and Tang, Ying and Lou, Yining and Chen, Yiru and Piao, Yishi and Chen, Yixuan and Xiong, Yong and Xuan, Yuchen and Yang, Yuehan and Xu, Yuer and Zha, Yukun and Ma, Yunxian and Lin, Yuping and Yan, Yuting and Xie, Yutong and Sheng, Yuwen and Zhu, Yuxuan and Zhang, Zekai and Ju, Zhe and Lin, Zhenzhen and Gao, Zheren and Sun, Zheyang and Yan, Zhigang and Wu, Zhongyu and Wang, Zi and Qu, Zihua and Yan, Ziling and Wan, Ziyi},
	year = {2026},
}

@misc{Qwen_3.8_flash_next,
    title  = {{Qwen3.8-Flash-Next}: A New Architecture, Towards Ultimate Cost-Efficiency},
    author = {{Qwen Team}},
    year   = {2026},
  note = {\url{https://qwen.ai/blog?id=qwen3.8-flash-next}}
}

@inproceedings{gpt_oss_120b,
      title={gpt-oss-120b \& gpt-oss-20b Model Card}, 
      booktitle = ARXIV,
      author = {OpenAI and Sandhini Agarwal and Lama Ahmad and Jason Ai and Sam Altman and Andy Applebaum and Edwin Arbus and Rahul K. Arora and Yu Bai and Bowen Baker and Haiming Bao and Boaz Barak and Ally Bennett and Tyler Bertao and Nivedita Brett and Eugene Brevdo and Greg Brockman and Sebastien Bubeck and Che Chang and Kai Chen and Mark Chen and Enoch Cheung and Aidan Clark and Dan Cook and Marat Dukhan and Casey Dvorak and Kevin Fives and Vlad Fomenko and Timur Garipov and Kristian Georgiev and Mia Glaese and Tarun Gogineni and Adam Goucher and Lukas Gross and Katia Gil Guzman and John Hallman and Jackie Hehir and Johannes Heidecke and Alec Helyar and Haitang Hu and Romain Huet and Jacob Huh and Saachi Jain and Zach Johnson and Chris Koch and Irina Kofman and Dominik Kundel and Jason Kwon and Volodymyr Kyrylov and Elaine Ya Le and Guillaume Leclerc and James Park Lennon and Scott Lessans and Mario Lezcano-Casado and Yuanzhi Li and Zhuohan Li and Ji Lin and Jordan Liss and Lily and Liu and Jiancheng Liu and Kevin Lu and Chris Lu and Zoran Martinovic and Lindsay McCallum and Josh McGrath and Scott McKinney and Aidan McLaughlin and Song Mei and Steve Mostovoy and Tong Mu and Gideon Myles and Alexander Neitz and Alex Nichol and Jakub Pachocki and Alex Paino and Dana Palmie and Ashley Pantuliano and Giambattista Parascandolo and Jongsoo Park and Leher Pathak and Carolina Paz and Ludovic Peran and Dmitry Pimenov and Michelle Pokrass and Elizabeth Proehl and Huida Qiu and Gaby Raila and Filippo Raso and Hongyu Ren and Kimmy Richardson and David Robinson and Bob Rotsted and Hadi Salman and Suvansh Sanjeev and Max Schwarzer and D. Sculley and Harshit Sikchi and Kendal Simon and Karan Singhal and Yang Song and Dane Stuckey and Zhiqing Sun and Philippe Tillet and Sam Toizer and Foivos Tsimpourlas and Nikhil Vyas and Eric Wallace and Xin Wang and Miles Wang and Olivia Watkins and Kevin Weil and Amy Wendling and Kevin Whinnery and Cedric Whitney and Hannah Wong and Lin Yang and Yu Yang and Michihiro Yasunaga and Kristen Ying and Wojciech Zaremba and Wenting Zhan and Cyril Zhang and Brian Zhang and Eddie Zhang and Shengjia Zhao},
      year={2025}
}

@inproceedings{vserve2026,
  title={{vServe}: Accurate {LLM} Serving Simulation for {PyTorch}'s {JIT}-Compiled Execution},
  author = {Bang, Jehyeon and Im, Jinseop and Chung, Jinha and Rhu, Minsoo},
  booktitle=ARXIV,
  year={2026}
}

@inproceedings{vtrain,
  title={{vTrain}: A Simulation Framework for Evaluating Cost-Effective and Compute-Optimal Large Language Model Training},
  author={Bang, Jehyeon and Choi, Yujeong and Kim, Myeongwoo and Kim, Yongdeok and Rhu, Minsoo},
  booktitle=MICRO,
  year={2024},
  note={\url{https://doi.org/10.1109/MICRO61859.2024.00021}}
}

@article{czitrom1999ofat,
  author={Czitrom, Veronica},
  title={One-Factor-at-a-Time versus Designed Experiments},
  journal={The American Statistician},
  volume={53},
  number={2},
  pages={126--131},
  year={1999}
}
